\documentclass[letterpaper, 10 pt, conference]{ieeeconf}  

\IEEEoverridecommandlockouts                              

\usepackage{epsfig} 
\usepackage{amsmath} 
\usepackage{amssymb}  
\usepackage{bm}  
\usepackage{subcaption}
\usepackage{siunitx}
\usepackage{hyperref}

\graphicspath{{Figures/}}

\title{\LARGE \bf
Analysis and Transfer of Human Movement Manipulability\\ in Industry-like Activities
}

\author{No\'{e}mie Jaquier, Leonel Rozo and Sylvain Calinon
	\thanks{N. Jaquier and S. Calinon are with the Idiap Research Institute, CH-1920 Martigny, Switzerland (e-mail: \tt\scriptsize firstname.lastname@idiap.ch).}
	\thanks{L. Rozo is with the Bosch Center for Artificial Intelligence (BCAI), 71272 Renningen, Germany (e-mail: \tt\scriptsize leonel.rozo@de.bosch.com).}
	\thanks{This work was partially supported by the TACT-HAND (SNSF/DFG) and COLLABORATE (EC H2020 Program, G.A. 820767) projects.}
	}

\newcommand{\etal}{\MakeLowercase{\textit{et al.\ }}}
\newcommand{\trsp}{\mathsf{T}}

\newcommand{\Sc}{\mathsf{Sc}} 
\newcommand{\Rec}{\mathsf{Re}} 
\newcommand{\Pic}{\mathsf{Pi}} 
\newcommand{\Ca}{\mathsf{Ca}} 
\newcommand{\Pl}{\mathsf{Pl}} 
\newcommand{\Fm}{\mathsf{Fm}} 
\newcommand{\Rl}{\mathsf{Rl}} 

\begin{document}

\maketitle
\thispagestyle{empty}
\pagestyle{empty}

\begin{abstract}
Humans exhibit outstanding learning, planning and adaptation capabilities while performing different types of industrial tasks. Given some knowledge about the task requirements, humans are able to plan their limbs motion in anticipation of the execution of specific skills. For example, when an operator needs to drill a hole on a surface, the posture of her limbs varies to guarantee a stable configuration that is compatible with the drilling task specifications, e.g. exerting a force orthogonal to the surface. Therefore, we are interested in analyzing the human arms motion patterns in industrial activities. To do so, we build our analysis on the so-called manipulability ellipsoid, which captures a posture-dependent ability to perform motion and exert forces along different task directions. Through thorough analysis of the human movement manipulability, we found that the ellipsoid shape is task dependent and often provides more information about the human motion than classical manipulability indices.
Moreover, we show how manipulability patterns can be transferred to robots by learning a probabilistic model and employing a manipulability tracking controller that acts on the task planning and execution according to predefined control hierarchies. 
\end{abstract}

\section{INTRODUCTION}
When performing manipulation tasks, we naturally put our limbs in a posture that best allows us to carry out the task at hand given specific workspace constraints. This posture adaptation alters the motion and strength characteristics of our arms so that they are compatible with specific task requirements. For example, the arm kinematics seems to play a central role when we plan point-to-point reaching movements, where joint trajectory patterns arise as a function of visual targets~\cite{Morasso81:SpatialControl}, indicating that task requirements lead to arm posture variations. This insight was also observed in more complex scenarios, where both kinematic and biomechanical factors affect task planning~\cite{Cos11:BiomechanicsInfluence}. For instance, Sabes and Jordan~\cite{Sabes97:MotorPlanning} observed that our central nervous system plans arm movements considering its directional sensitivity, which is directly related to the arm posture. This allows humans to be mechanically resistant to potential perturbations. Interestingly, directional preferences of human arm movements tend to exploit interaction torques for movement production at the shoulder or elbow, indicating that the preferred directions are largely determined by biomechanical factors~\cite{Dounskaia14:PreferDirections}. 

Roboticists have also investigated the impact of robot posture on manipulation tasks that involve pushing, pulling and reaching. It is well known that by varying the posture of a robot, we can change the optimal directions for motion generation or force exertion. This has direct implications in hybrid control, since the controller capability can be fully realized when the optimal directions for controlling velocity and force coincide with those dictated by the task~\cite{Chiu87:RedundantRbtCtrl}. In this context, the so-called manipulability ellipsoid~\cite{Yoshikawa85:Manipulabilty} serves as a geometric descriptor that, given a joint configuration, indicates the capability to arbitrarily perform motion and exert a force along the different task directions. 

Manipulability ellipsoids have been used to analyze the coordination of the human arm during reaching-to-grasp tasks for designing ergonomic environments~\cite{Jacquier-Bret12}, and to study the swing phase of human walking motion~\cite{MiripourFard19}. However, analyses of the human arm manipulability remain limited to few simple movements. Moreover, most of the conducted studies focus on the evolution of the manipulability volume and isotropy. In contrast, considering the direction of the major axis of manipulability ellipsoids has been proved useful in several human movement analysis works, notably in exoskeletons design and control. Goljat \etal~\cite{Goljat17} used the shape of the muscular manipulability of the human arm for controlling arm exoskeletons. They computed a varying support based on the main direction of the user's force manipulability. Inspired by human walking studies, Kim \etal~\cite{Kim10} proposed an energy-efficient gait pattern for leg exoskeletons which aligns the direction of motion with the major axis of the dynamic manipulability ellipsoid. While the major axis of the ellipsoid provides some information about the arm movement, the importance of the ellipsoid shape should not be neglected, as a low dexterity in motion along a specific axis is closely related to a high flexibility in force along the very same direction~\cite{Lee89:DualArmManipulability}. 

In this paper we analyze single and dual-arm manipulability of human movements during the execution of industry-like activities from a geometry-aware perspective. To do so, we use kinematic data records of several participants performing various activities such as screwing and load carrying~\cite{Maurice19:IndustryDataset}. Moreover, we consider an important characteristic of manipulability ellipsoids that was often overlooked in the literature, namely, the fact that they lie on the manifold of symmetric positive definite (SPD) matrices. We exploit differential geometry to statistically study the manipulability profile of human movements (see Section~\ref{sec:Bckgr} for a short background). The mean and variance of the ellipsoids provide more information about human motion than the classical manipulability indices related to the ellipsoids volume and isotropy, as explained in Section~\ref{sec:analysis}. Finally, we show that the observed task-dependent patterns can be transferred to robots as manipulability requirements when executing similar tasks, bypassing the complexity of kinematic mapping approaches (see Section~\ref{sec:transfer}).

The contributions of this paper are twofold: \emph{(i)} we provide a thorough statistical analysis of single and dual-arm manipulability of human movements in industry-like activities; \emph{(ii)} we demonstrate that the analyzed posture-dependent task requirements can be transferred from a human to robotic agents via a manipulability transfer framework. A video accompanying the paper is available at \url{https://youtu.be/q0GZwvwW9Ag}.

\section{Background}
\label{sec:Bckgr}

\subsection{Manipulability ellipsoids}
\label{subsec:ManEll}
Velocity and force manipulability ellipsoids introduced in~\cite{Yoshikawa85:Manipulabilty} are kinetostatic performance measures of robotic platforms. They indicate the preferred directions in which force or velocity control commands may be performed at a given joint configuration. More specifically, the velocity manipulability ellipsoid describes the characteristics of feasible Cartesian motion corresponding to all the unit norm joint velocities. The velocity manipulability of an $n$-DoF robot can be computed from the relationship between task velocities $\bm{\dot{x}}$ and joint velocities $\bm{\dot{q}}$, namely $\bm{\dot{x}} = \bm{J}(\bm{q}) \bm{\dot{q}}$, where $\bm{q}\!\in\!\mathbb{R}^n$ and $\bm{J}\!\in\!\mathbb{R}^{6 \times n}$ are the joint position and Jacobian of the robot, respectively. 
To do so, we consider a unit hypersphere $\|\bm{\dot{q}}\|^2 \!=\! 1$ in the joint velocity space, which is mapped into the Cartesian velocity space $\mathbb{R}^6$ with\footnote{Scaling of the joint velocities may be used to reflect actuator properties.}
\begin{align} 
\| \bm{\dot{q}} \|^2  = \bm{\dot{q}}^\trsp \bm{\dot{q}}  = \bm{\dot{x}}^\trsp(\bm{J}\bm{J}^\trsp)^{-1}\bm{\dot{x}},
\label{Eq:VelocityMapping}
\end{align}
by using the least-squares inverse kinematics solution ${\bm{\dot{q}}\!=\!\bm{J}^\dagger\bm{\dot{x}}\!=\!\bm{J}^\trsp(\bm{J}\bm{J}^\trsp)^{-1}\bm{\dot{x}}}$. Equation \eqref{Eq:VelocityMapping} represents the robot manipulability in terms of motion, indicating the flexibility of the manipulator in generating velocities in Cartesian space.
The major axis of the velocity manipulability ellipsoid $\bm{M}^{\bm{\dot{x}}} = \bm{J}\bm{J}^\trsp$, aligned to the eigenvector associated with the maximum eigenvalue $\lambda^{\tiny{\bm{M}^{\bm{\dot{x}}}}}_{\mathrm{max}}$, indicates the direction in which the greater velocity can be generated. That is in turn the direction along which the robot is more sensitive to perturbations. This occurs due to the principal axes of the force manipulability $\bm{M}^{\bm{F}}=(\bm{J}\bm{J}^\trsp)^{-1}$ being aligned with those of the velocity manipulability, with reciprocal lengths (eigenvalues) due to the velocity-force duality (see~\cite{Chiu87:RedundantRbtCtrl,Yoshikawa85:Manipulabilty} for details). 

Other forms of manipulability ellipsoids exist, such as the dynamic manipulability~\cite{Yoshikawa85:DinamicMan}, which gives a measure of the ability of performing end-effector accelerations in the task space for a given set of joint torques. 
As another example, the muscular manipulability~\cite{Goljat17} indicates the end-effector forces that can be generated in function of the muscle forces.

As mentioned previously, any manipulability ellipsoid $\bm{M}$ belongs to the set of SPD matrices $\mathcal{S}_{++}^D$ which describe the interior of a convex cone. Consequently, we must consider this particular aspect to properly analyze manipulability profiles. The corresponding Riemannian manifold and the computation of manipulability statistics are introduced next.  

\subsection{Riemannian manifold of SPD matrices}
\label{subsec:Riemannian}
The set of $D\!\times\!D$ SPD matrices $\mathcal{S}_{++}^D$ is not a vector space since it is not closed under addition and scalar product, but instead forms a Riemannian manifold~\cite{Pennec06}. A $d$-dimensional manifold $\mathcal{M}$ is a topological space which is locally Euclidean, which means that each point in $\mathcal{M}$ has a neighborhood which is homeomorphic to an open subset of the $d$-dimensional Euclidean space $\mathbb{R}^d$. A Riemannian manifold $\mathcal{M}$ is a differentiable manifold equipped with a Riemannian metric. 

For each point $\bm{\Sigma}\!\in\!\mathcal{M}$, there exists a tangent space $\mathcal{T}_{\bm{\Sigma}} \mathcal{M}$ which is formed by the tangent vectors to all $1$-dimensional curves on $\mathcal{M}$ passing through $\bm{\Sigma}$. The origin of the tangent space coincides with $\bm{\Sigma}$. In the case of the SPD manifold, the tangent space at any point ${\bm{\Sigma}\in\mathcal{S}_{++}^D}$ corresponds to the space of symmetric matrices $\text{Sym}^D$ (see Fig.~\ref{Fig:SPD}-\emph{left}). The Riemannian metric is a smoothly-varying positive-definite inner product $\langle \cdot, \cdot \rangle_{\bm{\bm{\Sigma}}}: \mathcal{T}_{\bm{\bm{\Sigma}}}\mathcal{M} \times \mathcal{T}_{\bm{\bm{\Sigma}}} \mathcal{M}\to \mathcal{M}$ acting on $\mathcal{T}_{\bm{\bm{\Sigma}}} \mathcal{M}$. 

The Riemannian metric allows us to define the Riemannian distance between two points $\bm{\Sigma},\bm{\Lambda}\in\mathcal{M}$, which corresponds to the minimum length over all possible smooth curves on the manifold between $\bm{\Sigma}$ and $\bm{\Lambda}$.
The corresponding curve is called a \emph{geodesic}. Geodesics are the generalization of straight lines to Riemannian manifolds, as they are locally length-minimizing curves with constant speed in $\mathcal{M}$. 

As a Riemannian manifold is not a vector space, the use of classical Euclidean space methods for treating and analyzing data lying on this manifold is inadequate. However, due to their Euclidean geometry, linear algebra operations can be performed on the elements of each tangent space of a manifold. To utilize these Euclidean tangent spaces, we need mappings back and forth between $\mathcal{T}_{\bm{\Sigma}} \mathcal{M}$ and $\mathcal{M}$, which are known as \emph{exponential} and \emph{logarithmic maps}.
The exponential map $\text{Exp}_{\bm{\Sigma}}: \mathcal{T}_{\bm{\Sigma}} \mathcal{M}\to \mathcal{M}$ maps a point $\bm{L}$ in the tangent space of $\bm{\Sigma}$ to a point $\bm{\Lambda}$ on the manifold, so that it is reached at time $1$ by the geodesic starting at $\bm{\Sigma}$ in the direction $\bm{L}$, i.e., $d_{\mathcal{M}}(\bm{\Sigma},\bm{\Lambda}) = \|\bm{L}\|_{\bm{\Sigma}}$. The inverse operation is called the logarithmic map $\text{Log}_{\bm{\Sigma}}:  \mathcal{M}\to \mathcal{T}_{\bm{\Sigma}}\mathcal{M}$.
Specifically, the exponential and logarithmic maps on the SPD manifold corresponding to the affine-invariant distance 
\begin{equation}
d_{\mathcal{S}_{++}^D}(\bm{\Lambda},\bm{\Sigma}) = \|\log(\bm{\Sigma}^{-\frac{1}{2}}\bm{\Lambda}\bm{\Sigma}^{-\frac{1}{2}})\|_\text{F},
\label{Eq:SPDdist}
\end{equation}
are computed as (see~\cite{Pennec06} for details)
\begin{align}
\bm{\Lambda} & = \text{Exp}_{\bm{\Sigma}}(\bm{L}) = \bm{\Sigma}^{\frac{1}{2}}\exp(\bm{\Sigma}^{-\frac{1}{2}}\bm{L}\bm{\Sigma}^{-\frac{1}{2}})\bm{\Sigma}^{\frac{1}{2}}, \\
\bm{L} & = \text{Log}_{\bm{\Sigma}}(\bm{\Lambda}) = \bm{\Sigma}^{\frac{1}{2}}\log(\bm{\Sigma}^{-\frac{1}{2}}\bm{\Lambda}\bm{\Sigma}^{-\frac{1}{2}})\bm{\Sigma}^{\frac{1}{2}},
\label{Eq:SPDmaps}
\end{align}
where $\exp(\cdot)$ and $\log(\cdot)$ are matrix functions. These operations are illustrated in Fig.~\ref{Fig:SPD}-\emph{left}. 

In this paper, we first exploit the Riemannian manifold framework to compute statistics, such as the mean and covariance, of manipulability ellipsoids profiles. A Riemannian treatment is necessary to ensure that the mean of manipulability ellipsoids is valid and unique, meaning that it belongs to the space of SPD matrices.
The Fr\'echet mean~\cite{Frechet48:ManifoldMean} of a set of $N$ datapoints $\bm{\Sigma}_n\in\mathcal{S}_{++}^D$ corresponds to the matrix $\bm{\Xi}\in\mathcal{S}_{++}^D$ that minimizes the sum of the squared affine-invariant distances $\sum_{n=1}^{N}d_{\mathcal{S}_{++}^D}(\bm{\Xi},\bm{\Sigma}_n)$, as detailed in~\cite{Pennec06}. This optimization problem can be solved iteratively with a Gauss-Newton algorithm. At each iteration, the datapoints are first projected into the tangent space of the current estimate of the mean $\bm{\Xi}$ using the logarithmic map. Then, the Euclidean mean of these points is computed and projected to the manifold using the exponential map, which corresponds to the updated estimate of the mean, i.e.,
\begin{equation}
\bm{\Xi} \gets \text{Exp}_{\bm{\Xi}}\!\left(\frac{1}{N}\sum_{n=1}^{N}\;\text{Log}_{\bm{\Xi}}\!(\bm{\Sigma}_n)\right).
\label{Eq:SPDmean}
\end{equation}
The covariance tensor $\bm{\mathcal{S}}\in\mathbb{R}^{D\times D\times D\times D}$ is then computed in the tangent space of the mean as
\begin{equation}
\bm{\mathcal{S}} = \frac{1}{N-1}\sum_{n=1}^{N}\text{Log}_{\bm{\Xi}}\!(\bm{\Sigma}_n)\otimes \text{Log}_{\bm{\Xi}}\!(\bm{\Sigma}_n),
\label{Eq:SPDcov}
\end{equation}
where $\otimes$ represents the tensor product. 
The tensor product of two matrices $\bm{X}\in \mathbb{R}^{I_1\times I_2}$, $ \bm{Y}\in \mathbb{R}^{J_1\times J_2}$ is  $\bm{X}\otimes\bm{Y} \in \mathbb{R}^{I_1\times I_2\times J_1\times J_2}$ with elements ${(\bm{X}\otimes\bm{Y})_{i_1,i_2,j_1,j_2} \; = \; x_{i_1,i_2} \; y_{j_1,j_2}}$. The concepts of mean and covariance on the SPD manifold $\mathcal{S}_{++}^2$ are illustrated in Fig.~\ref{Fig:SPD}-\emph{right}.

The Riemannian manifold framework is also exploited for human-robot manipulability transfer in Section~\ref{sec:transfer}. As shown in~\cite{Jaquier20}, a geometry-aware approach proves to be crucial for transferring manipulability requirements to robots in terms of accuracy, stability and convergence, beyond providing an appropriate mathematical treatment of the problem.

\newsavebox{\spdmat}
\savebox{\spdmat}{$\left(\begin{smallmatrix}T_{11} & T_{12}\\ T_{12} & T_{22}\end{smallmatrix}\right)$}
\begin{figure}[tbp]
	\centering
	\includegraphics[width=0.25\textwidth]{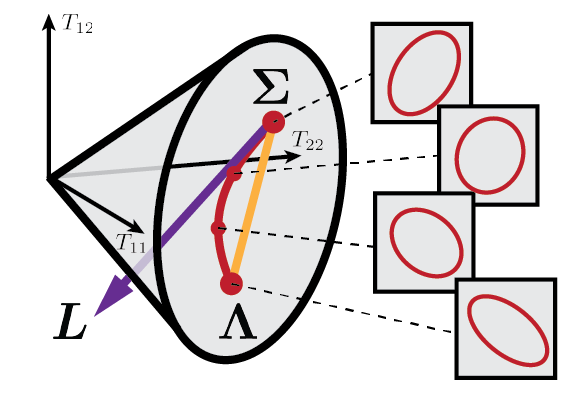}
	\includegraphics[width=.16\textwidth]{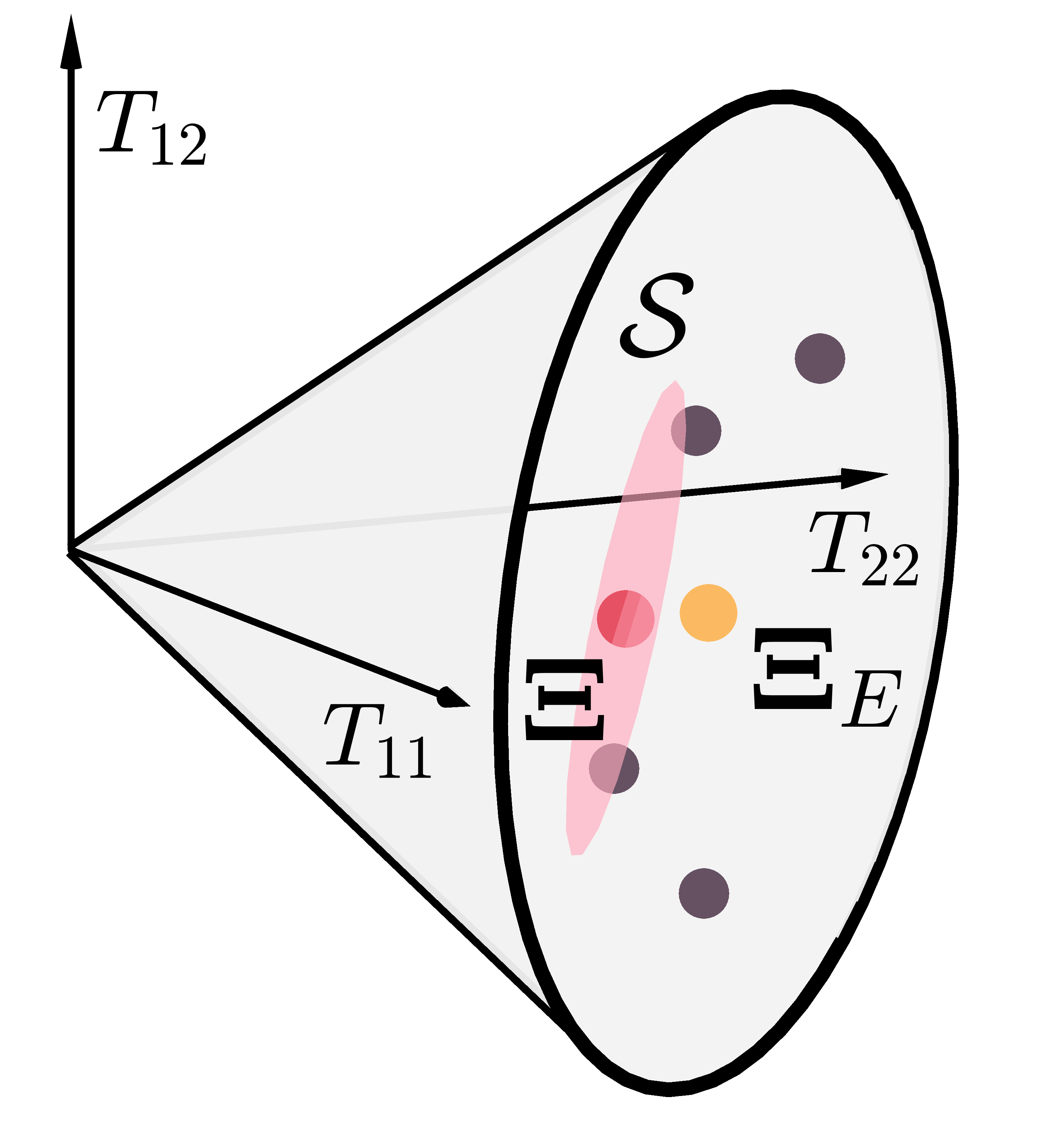}
	\caption{SPD manifold $\mathcal{S}_{++}^2$ embedded in its tangent space $\text{Sym}^2$. One point represents a matrix~\usebox{\spdmat} $\in \text{Sym}^2$. Points inside the cone, such as $\bm{\Sigma}$ and $\bm{\Lambda}$, belong to $\mathcal{S}_{++}^2$. \textbf{Left}: $\bm{L}$ lies on the tangent space of $\bm{\Sigma}$ such that $\bm{L} = \text{Log}_{\bm{\Sigma}}(\bm{\Lambda})$. The shortest path between $\bm{\Sigma}$ and $\bm{\Lambda}$ is the geodesic depicted as a red curve, which differs from the Euclidean path shown in yellow. \textbf{Right}: The Karcher mean $\bm{\Xi}$ and covariance tensor $\bm{\mathcal{S}}$ for the 4 points on the manifold are displayed as red point and ellipsoid, respectively. The mean $\bm{\Xi}$ differs from the Euclidean mean $\bm{\Xi}_E$ shown in yellow.}
	\label{Fig:SPD}
	\vspace{-0.45cm}
\end{figure}

\section{Manipulability Analysis}
\label{sec:analysis}
Analyzing the manipulability of the human arms during various tasks may be relevant to define desired manipulability ellipsoids of robots. The manipulability profile of a user while performing a task may provide relevant information about task planning and motion generation. For example, the main axis of the ellipsoid may indicate future directions of motion, while a small manipulability may reveal a lack of velocity or force control of an operator along specific directions. These aspects may notably be exploited to better design and control exoskeletons and ergonomic devices. In this section, we propose a detailed analysis of human arm(s) manipulability in industry-like activities by exploiting the mathematical concepts presented in Section~\ref{sec:Bckgr}. 

\subsection{Data description}
For our analysis, we use the industry-oriented dataset presented in~\cite{Maurice19:IndustryDataset}, which contains the whole-body posture data of 13 participants executing various industry-related activities. Each participant executed the tasks in 5 consecutive trials for 3 predefined sequences. In this paper, we consider 3 screwing motions and 2 carrying tasks provided in the database. A screwing task consists of taking a screw and a bolt from a \SI{75}{\centi\metre}-high table, walking to a shelf and screwing (with bare hands) at a specific height. The screwing movements realized at heights of \SI{60}{\centi\metre}, \SI{115}{\centi\metre} and \SI{175}{\centi\metre} are denoted by $\mathsf{SL}$, $\mathsf{SM}$ and $\mathsf{SH}$ for screw-low, -middle and -high, respectively. A carrying task involves taking a load from a \SI{55}{\centi\metre}-high table, walking to a shelf and put the load on it. Loads of \SI{5}{\kilogram} and \SI{10}{\kilogram} are placed in shelves of \SI{20}{\centi\metre} and \SI{110}{\centi\metre} high, respectively. The corresponding tasks are denoted by $\mathsf{C5}$ and $\mathsf{C10}$. The participants freely adapt their posture for each activity, as no explicit instructions were given for task execution. Therefore, we expect the arms manipulability to reflect the features of the natural motion of the participants. 

The labels of the motions in all the trials are provided along with the dataset, including the general and detailed posture, as well as the current action labeled by 3 independent annotators. In this paper, we study the human arm(s) manipulability according to the current action for the aforementioned carrying and screwing tasks. For each trial, we first identify the frames corresponding to each activity based on the labels and the order of activities defined in the corresponding sequence. Then, we select the frames corresponding to a subset of actions for each activity. The idea is to consider the motions corresponding to the relevant activities that are reproduced by all the participants. Therefore, we do not examine particular cases, e.g. when a screw falls on the ground and the participant needs to pick it up. The subset of actions considered for a screwing motion is composed of reaching ($\Rec$), picking ($\Pic$), carrying ($\Ca$), placing ($\Pl$), fine manipulation ($\Fm$), screwing ($\Sc$) and releasing ($\Rl$) the screw and the bolt. Concerning the carrying motion, we analyze the subset of actions composed of picking ($\Pic$), carrying ($\Ca$) and placing ($\Pl$) the load. 

The single- and dual-arm manipulability ellipsoids are computed for each time step of the different actions of the screwing and carrying tasks for 15 trials of 13 participants. The computation of the manipulability ellipsoids from the whole-body position and orientation data is described next.

\subsection{Human manipulability computation}
As the manipulability is a function of the Jacobian, we need a kinematic model of the human arm to compute its manipulability ellipsoid. In this paper we use the identification method for anthropomorphic arm kinematics proposed in~\cite{Ding13,Fang13}. The typical anthropomorphic arm model, where the arm is regarded as a 7-DoF manipulator, is exploited here along with the concept of human arm triangle, the latter used for joint-to-task space mapping. Thus, the joint angles and the corresponding Jacobian can be computed from the position and orientation of the wrist in task space, which in our case is given in the database. 

\begin{figure}[tbp]
	\centering
	\includegraphics[width=0.2\textwidth]{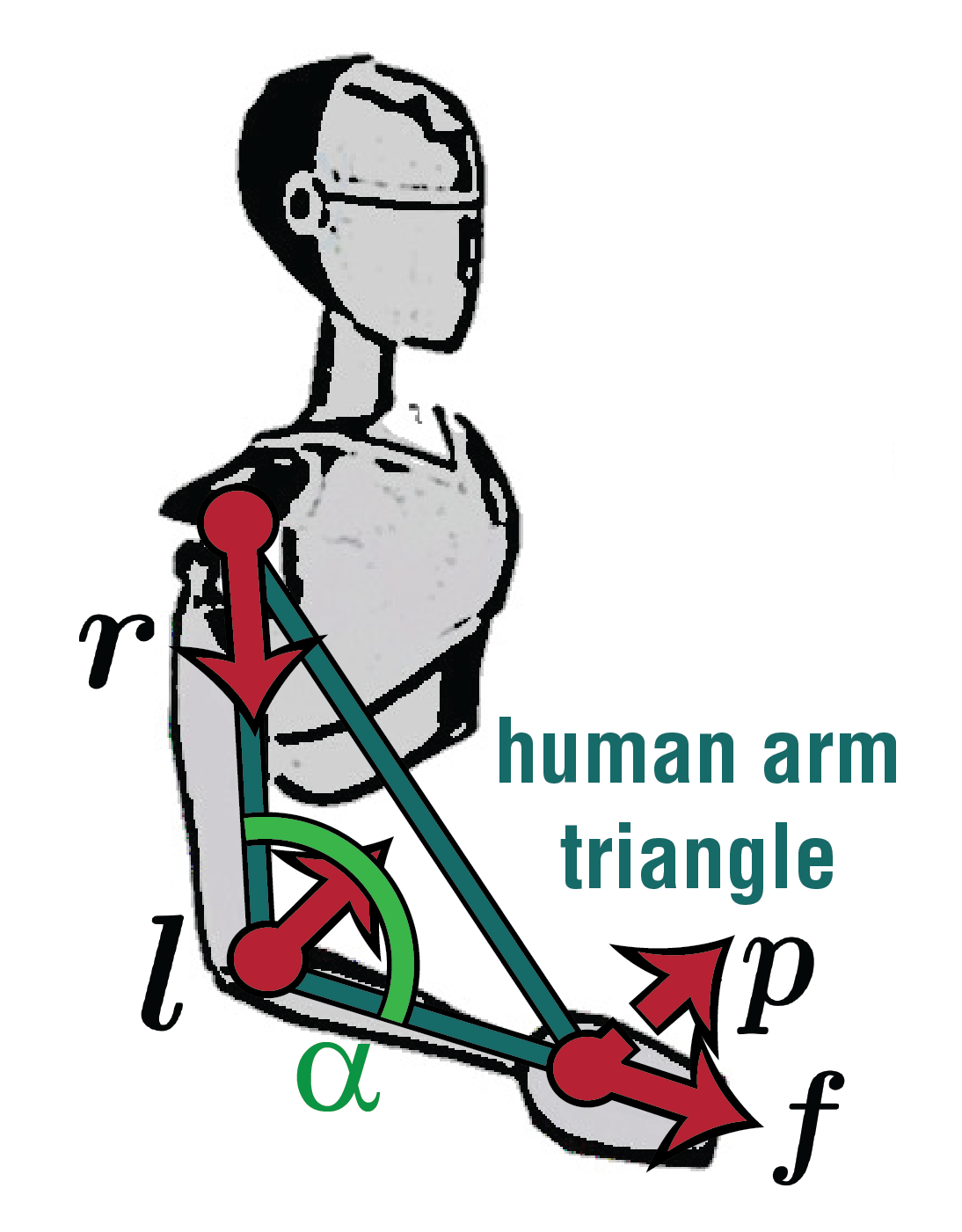}
	\caption{Human arm triangle model of 5 parameters. The triangle vertex are located at the center of the shoulder, elbow and wrist joints.}
	\vspace{-0.5cm}
	\label{Fig:HumanArmTriangle}
\end{figure}

Specifically, the human arm triangle model, shown in Fig.~\ref{Fig:HumanArmTriangle}, is defined by 5 parameters: the unit direction vector of the upper arm $\bm{r}$, the unit normal vector of the human arm triangle space $\bm{l}$, the angle between the upper and lower arm $\alpha$, the unit normal vector of the plane of the palm (pointing outward the palm) $\bm{p}$, and the unit direction vector of the fingers $\bm{f}$. The parameters $\{\bm{r}, \bm{l}, \alpha\}$ can be inferred from the wrist position in task space (see~\cite{Ding13} for details), while the parameters $\{\bm{p}, \bm{f}\}$ directly represent the wrist orientation. Moreover, it has been shown that the space spanned by the set of parameters $\{\bm{r}, \bm{l}, \alpha, \bm{p}, \bm{f}\}$ has a one-to-one relation with the joint space spanned by the seven joints $\{q_1 \ldots q_7\}$ of the anthropomorphic arm model. The formulas for the mappings $\{\bm{r}, \bm{l}, \alpha\} \to \{q_1 \ldots q_4\}$ and $\{\bm{p}, \bm{f}\} \to \{q_5 \ldots q_7\}$ are given in~\cite{Ding13} and~\cite{Fang13}, respectively.

This model allows us to compute the human arm Jacobian $\bm{J}$. The arm velocity manipulability ellipsoid is then computed as $\bm{M}^{\bm{\dot{x}}} = \bm{J}\bm{J}^\trsp$. For dual-arm manipulation tasks, such as carrying, we are interested in the manipulability ellipsoid of the dual-arm system. In this case, the set of joint velocities of constant unit norm $\|\bm{\dot{q}}_d\| = \|(\bm{\dot{q}}_l^\trsp, \bm{\dot{q}}_r^\trsp)^\trsp\| = 1$ is mapped to the Cartesian velocity space $\bm{\dot{x}}_d = (\bm{\dot{x}}_l^\trsp, \bm{\dot{x}}_r^\trsp)^\trsp$ through
\begin{equation} 
\| \bm{\dot{q}}_d \|^2  = \bm{\dot{q}}_d^\trsp \bm{\dot{q}}_d  = \bm{\dot{x}}_d^\trsp(\bm{G}_d^{\dagger\trsp}\bm{J}_d\bm{J}_d^\trsp\bm{G}_d^{\dagger})^{-1}\bm{\dot{x}}_d,
\label{Eq:MultipleArmsVelocityMapping}
\end{equation}
with Jacobian $\bm{J}_d = \text{diag}(\bm{J}_l, \bm{J}_r)$, grasp matrix $\bm{G}_d = (\bm{G}_l, \bm{G}_r)$ and indices $l$ and $r$ denoting the left and right arm, respectively. Therefore, the dual-arm velocity manipulability is given by $\bm{M}^{\bm{\dot{x}}}_d = \bm{G}_d^{\dagger\trsp}\bm{J}_d\bm{J}_d^\trsp\bm{G}_d^{\dagger}$~\cite{Chiacchio91}. Note that we assume two independent kinematic chains for the arms in the computation of $\bm{J}_d$. Moreover, the system is modeled under the assumption that the arms are holding a rigid object with a tight grasp. The force manipulability ellipsoid is the inverse of the velocity manipulability, i.e. $\bm{M}^{\bm{F}} = (\bm{M}^{\bm{\dot{x}}})^{-1}$.

\subsection{Analysis}
\begin{figure}[tbp]
	\centering
	\begin{subfigure}[b]{0.15\textwidth}
		\includegraphics[width=\textwidth]{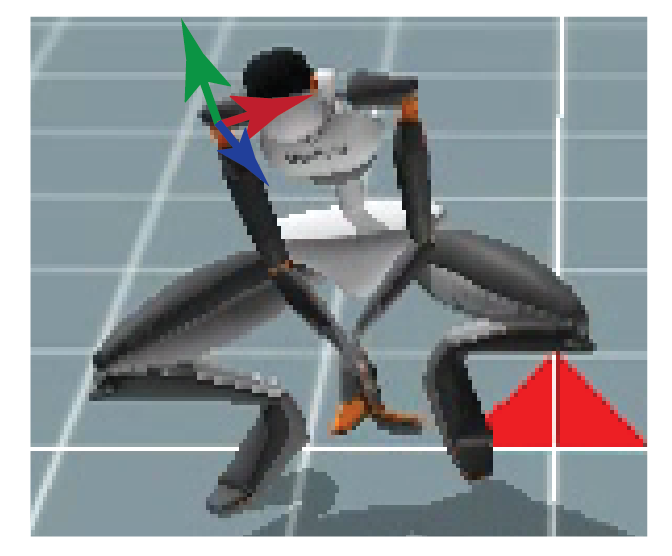}
		\includegraphics[width=\textwidth]{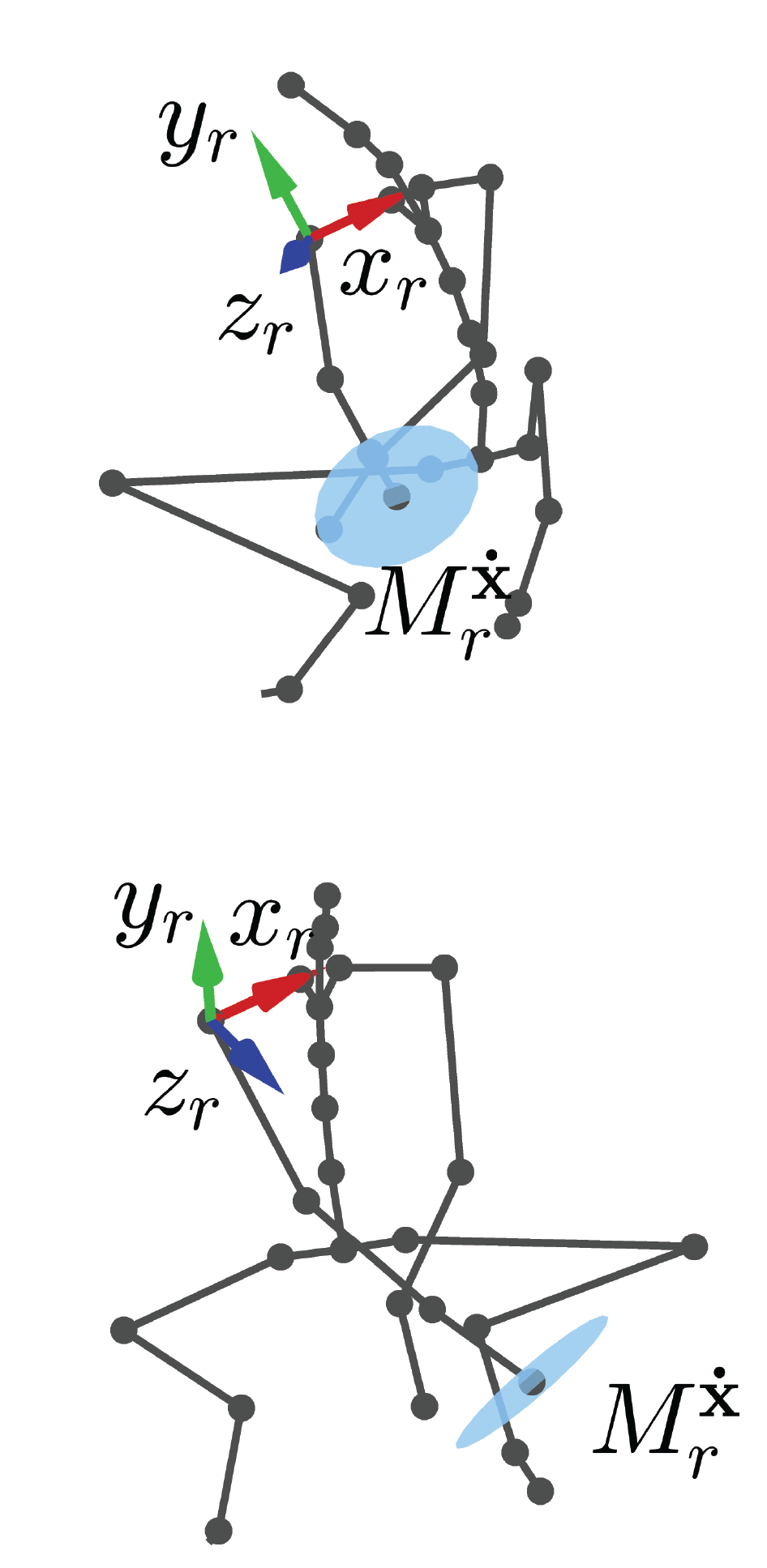}
		\caption{$\mathsf{SL}$}
		\label{subFig:SL}
	\end{subfigure}
	\begin{subfigure}[b]{0.15\textwidth}
		\includegraphics[width=\textwidth]{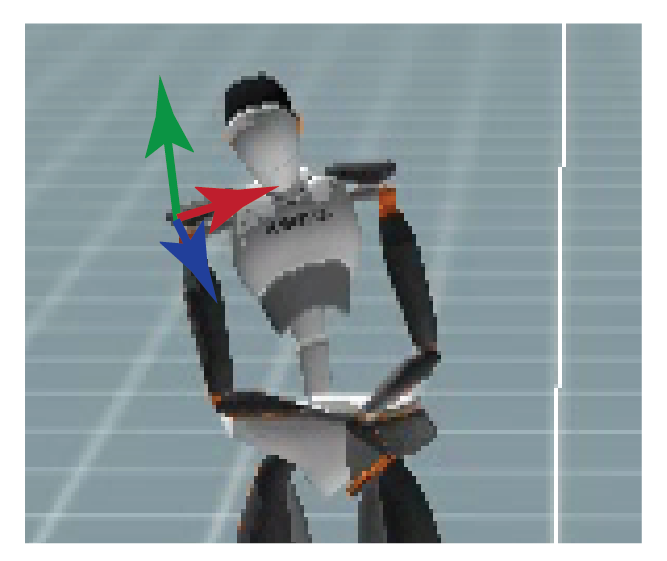}
		\includegraphics[width=\textwidth]{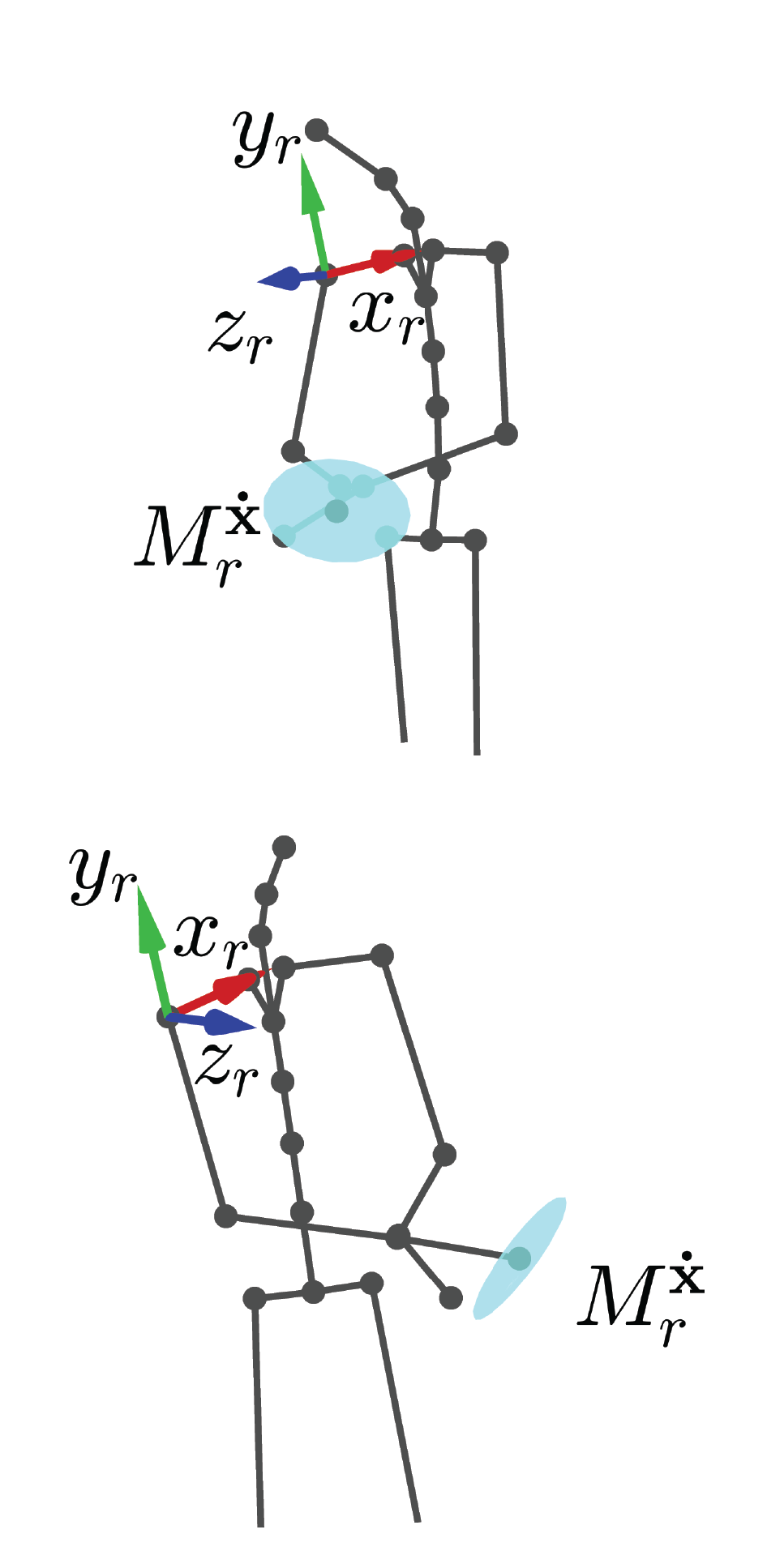}
		\caption{$\mathsf{SM}$}
		\label{subFig:SM}
	\end{subfigure}
	\begin{subfigure}[b]{0.15\textwidth}
		\includegraphics[width=\textwidth]{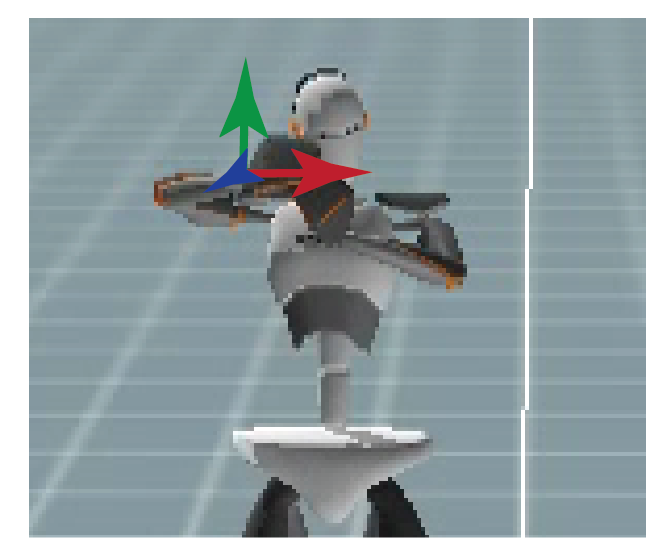}
		\includegraphics[width=\textwidth]{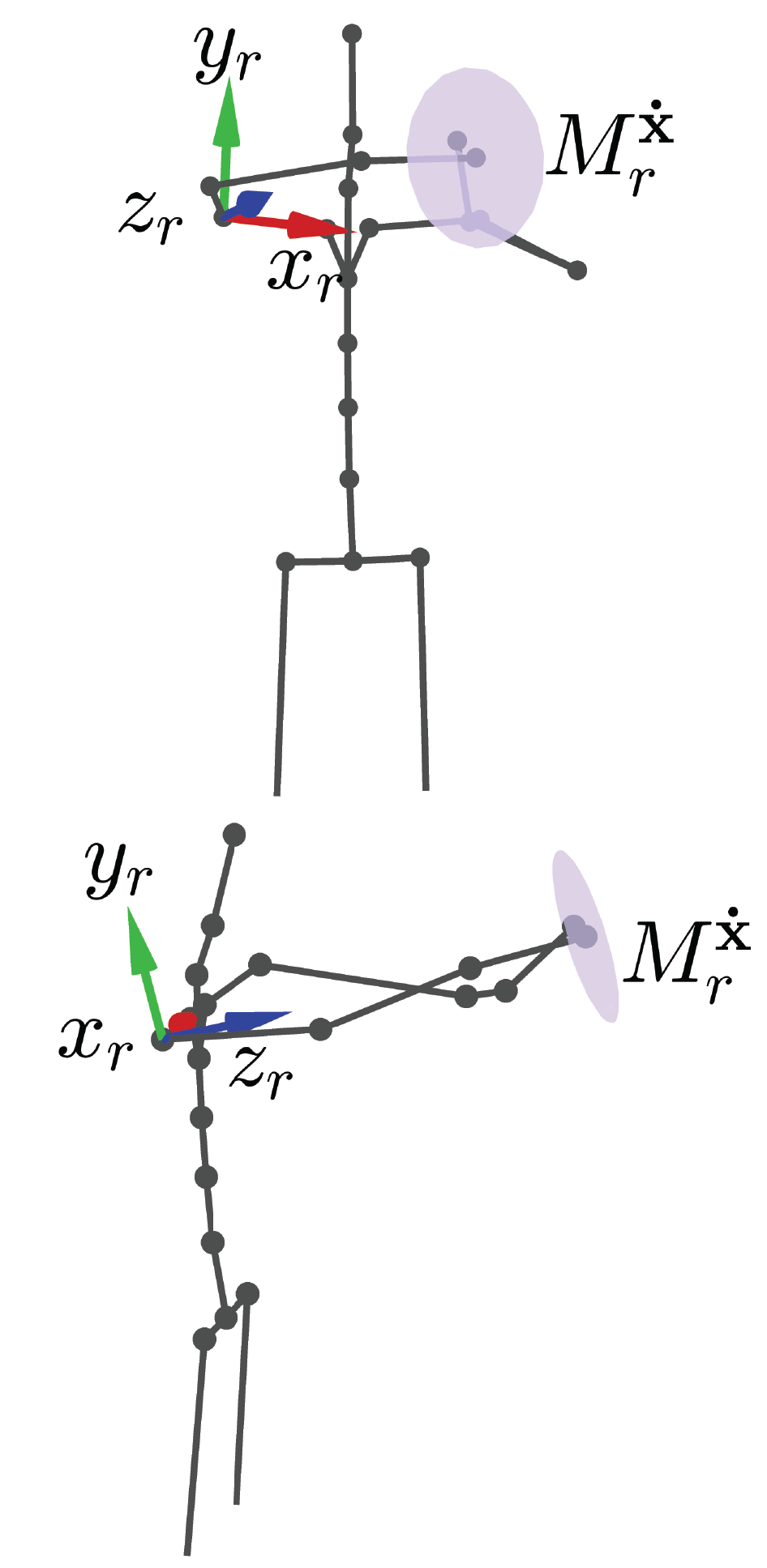}
		\caption{$\mathsf{SH}$}
		\label{subFig:SH}
	\end{subfigure}
	\caption{Posture of the participant 541 during the screwing actions at 3 different heights. The \emph{top} row shows the Xsens avatar view, while the \emph{middle} and \emph{bottom} rows display the skeleton model and right-arm velocity manipulability ellipsoid from 2 different viewpoints, along with the right shoulder reference frame. The manipulability ellipsoids are scaled by a factor $3$ for visualization purposes.}
	\label{Fig:HumanScrewing}
	\vspace{-.35cm}
\end{figure}

For screwing motions, we study the single-arm velocity manipulability ellipsoids. Figure~\ref{Fig:HumanScrewing} shows the posture of a participant while executing this task at different heights, along with the corresponding right-arm velocity manipulability ellipsoid. We observe that the shape of the ellipsoids is similar for the three screwing motions, regardless of the specified height. Namely, the ellipsoids shrink along the hand axis and isotropically elongate along the other directions. This indicates a high precision along the hand axis coupled with a high capability of motion on the orthogonal plane, where the hand is moving to execute the rotative screwing motion. This shows that the human arm manipulability is being adapted to the task requirements. 

\begin{figure*}[tbp]
	\centering
	\begin{subfigure}[b]{0.32\textwidth}
		\includegraphics[width=\textwidth]{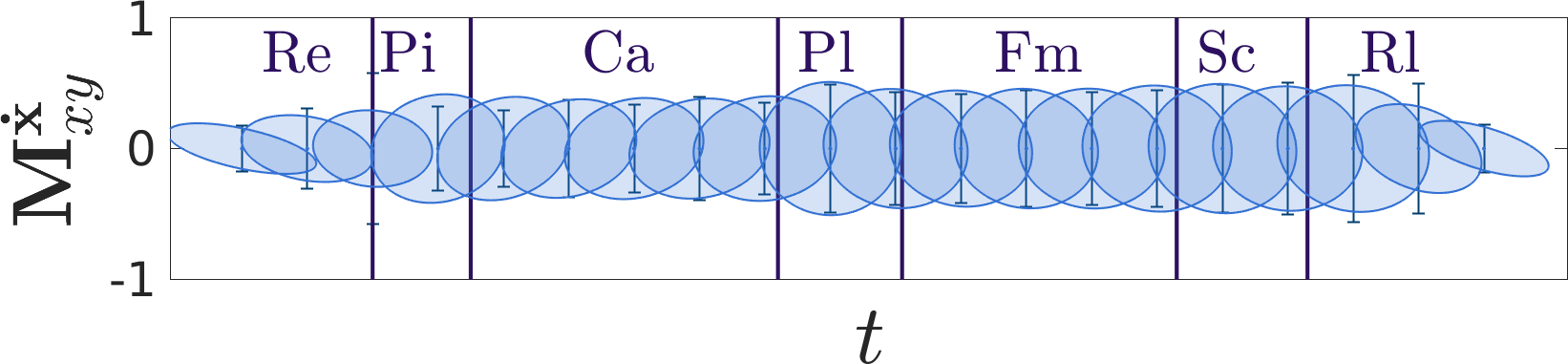}
		
		\vspace{0.1cm}
		\includegraphics[width=\textwidth]{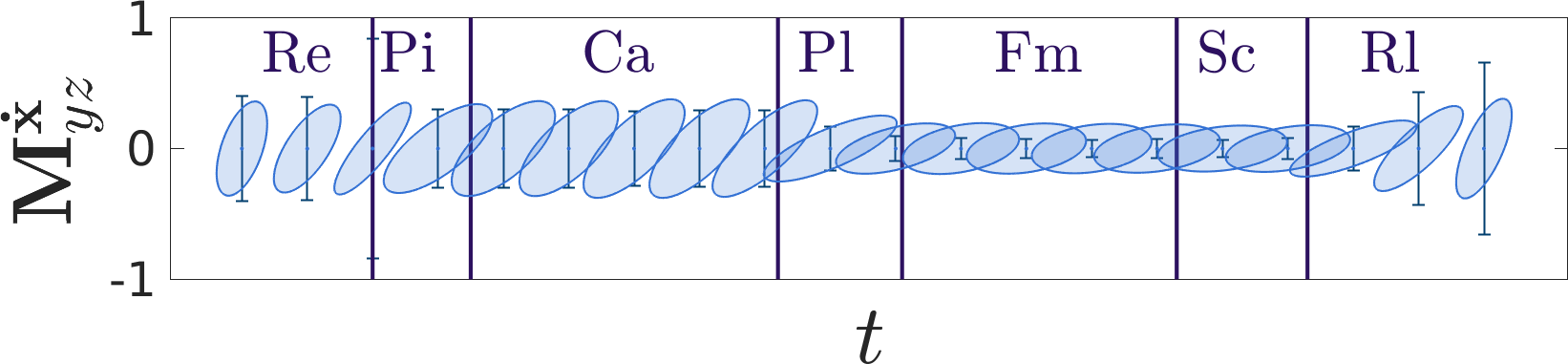}
		
		\vspace{0.1cm}
		\includegraphics[width=\textwidth]{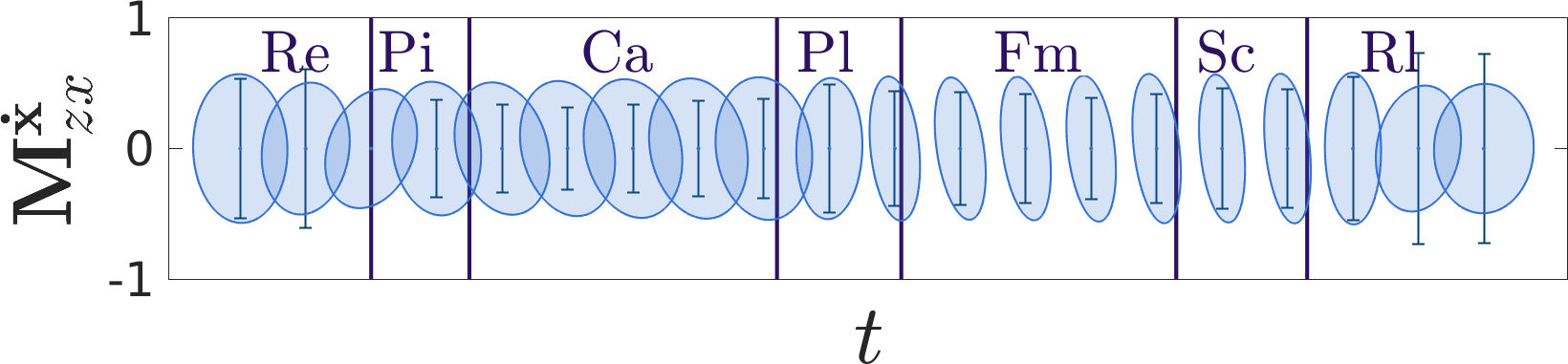}
		
		\vspace{0.25cm}
		\includegraphics[width=\textwidth]{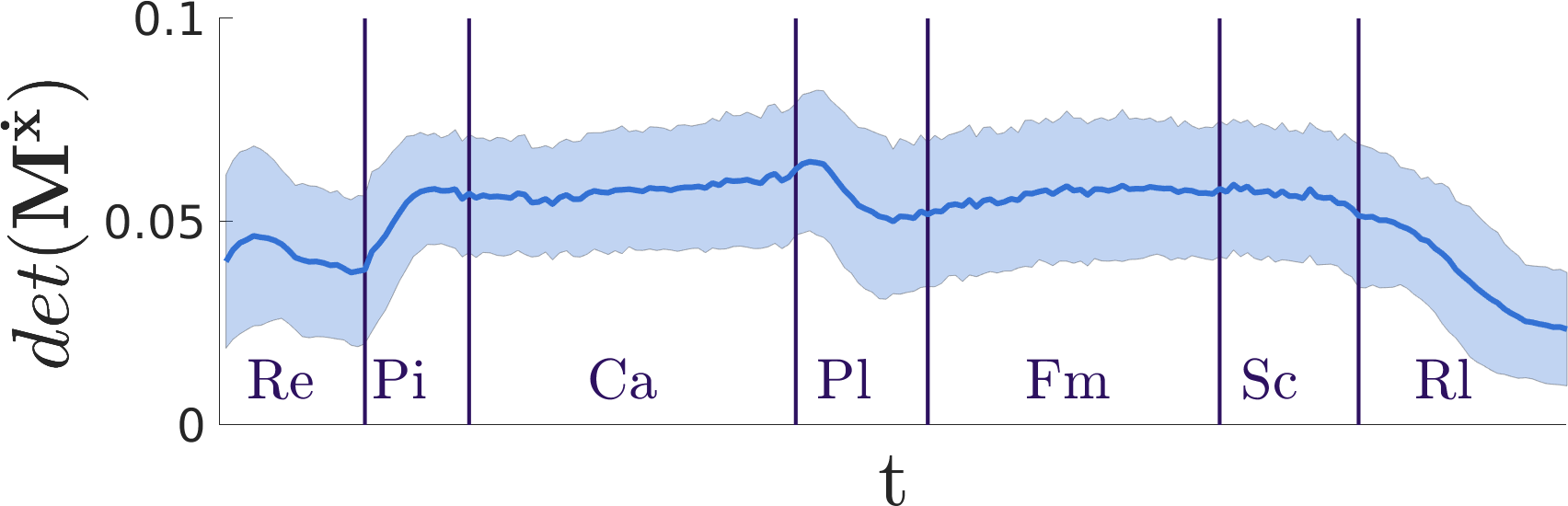}
%
		\includegraphics[width=\textwidth]{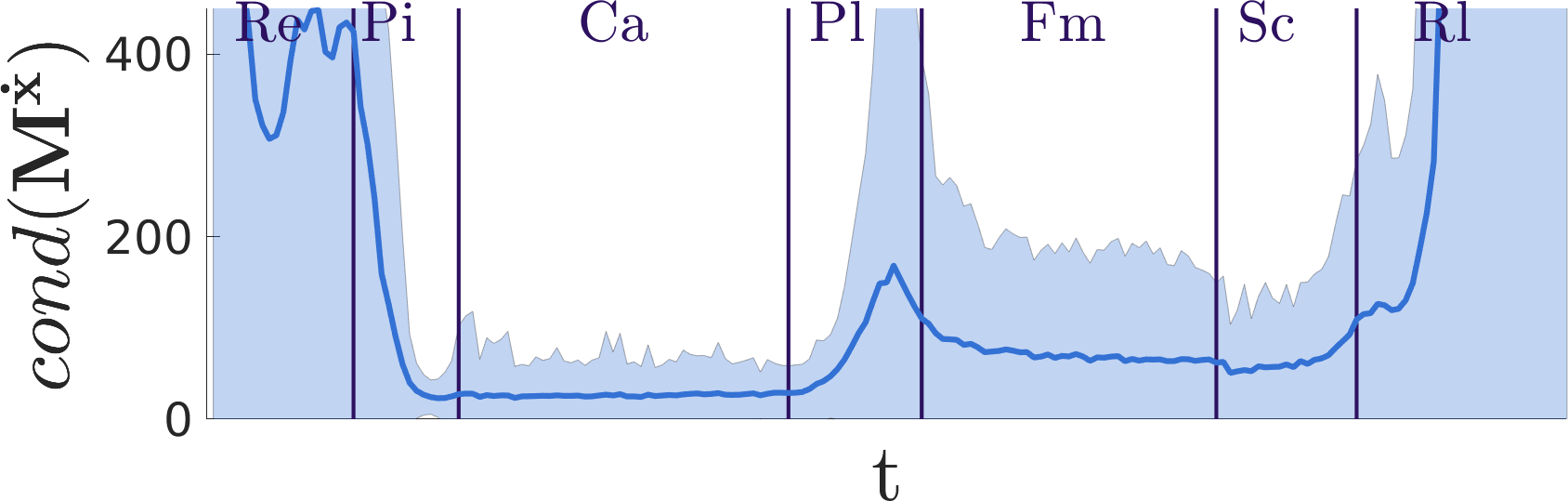}
		\caption{Screw low $\mathsf{SL}$}
		\label{subFig:SLall}
	\end{subfigure}
	\begin{subfigure}[b]{0.32\textwidth}
		\includegraphics[width=\textwidth]{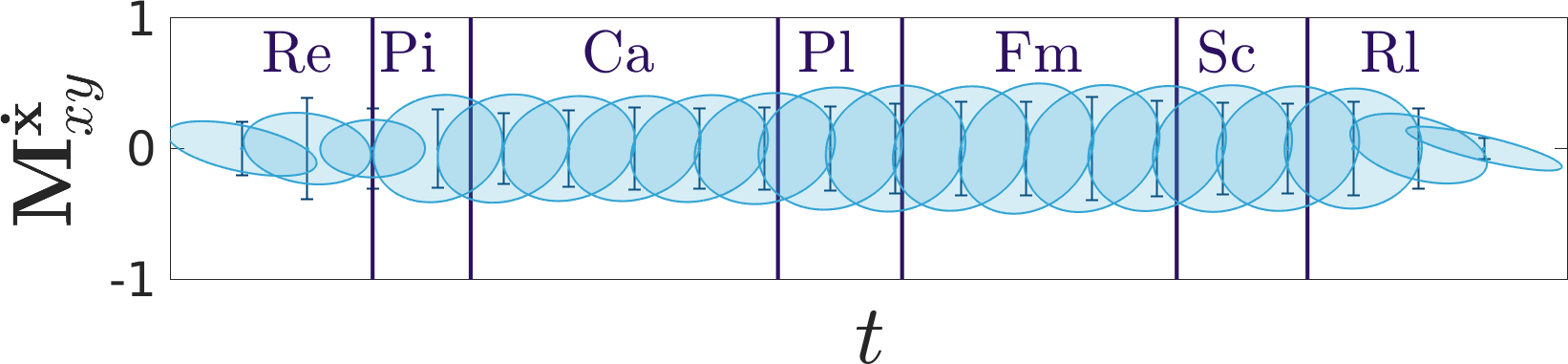}
		
		\vspace{0.1cm}
		\includegraphics[width=\textwidth]{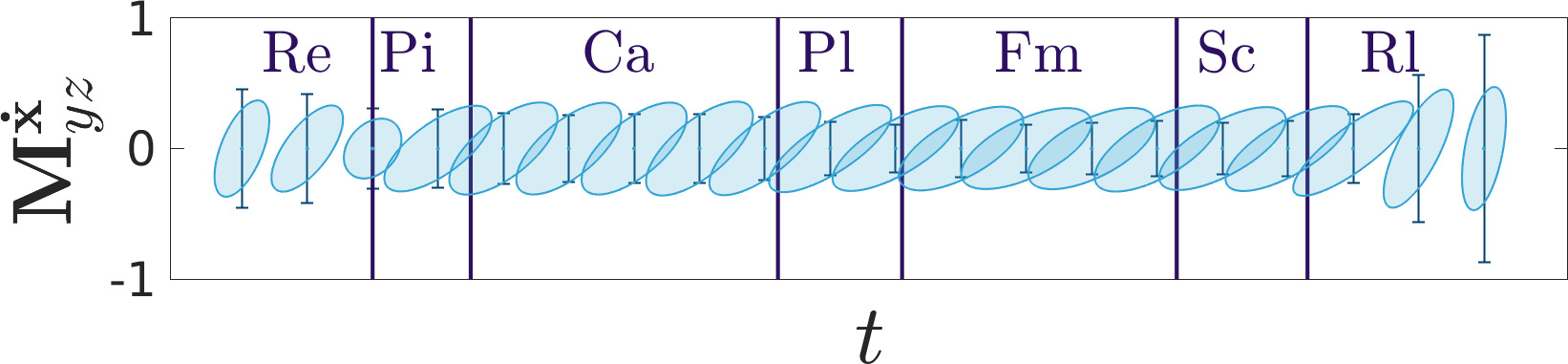}
		
		\vspace{0.1cm}
		\includegraphics[width=\textwidth]{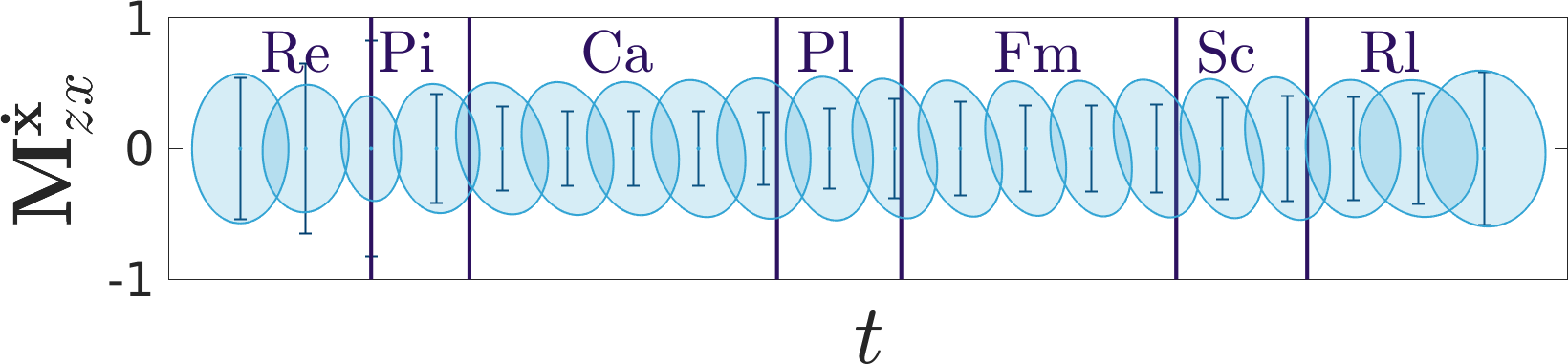}
		
		\vspace{0.1cm}
		\includegraphics[width=\textwidth]{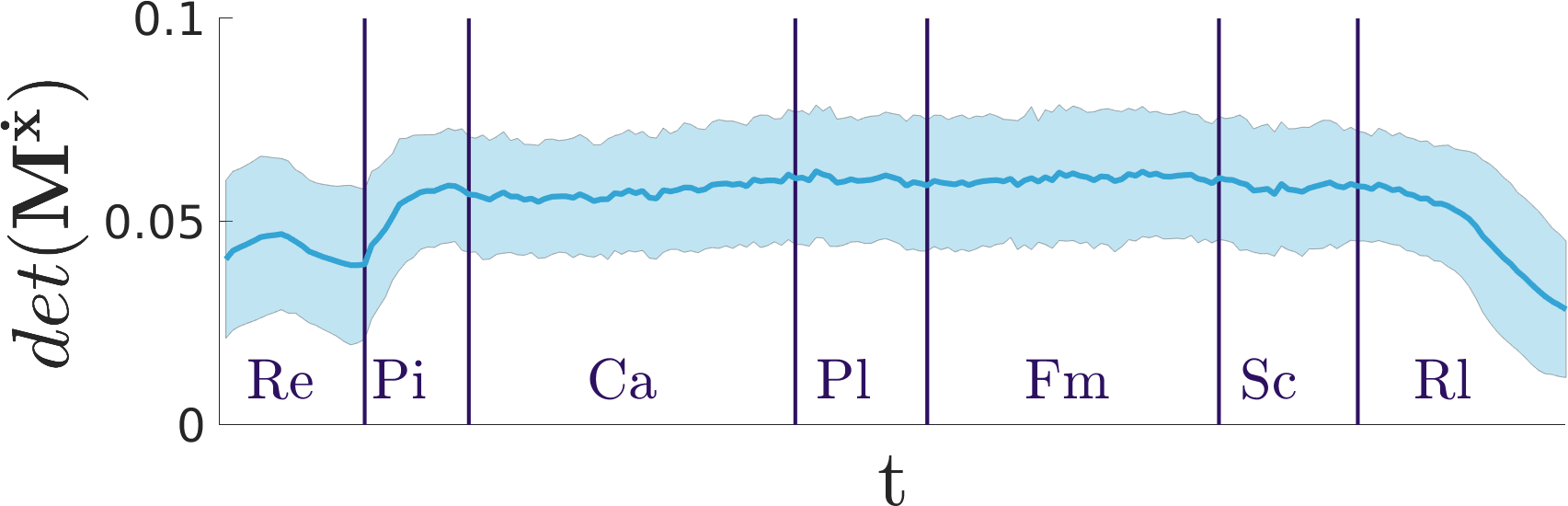}
		
		\vspace{0.1cm}
		\includegraphics[width=\textwidth]{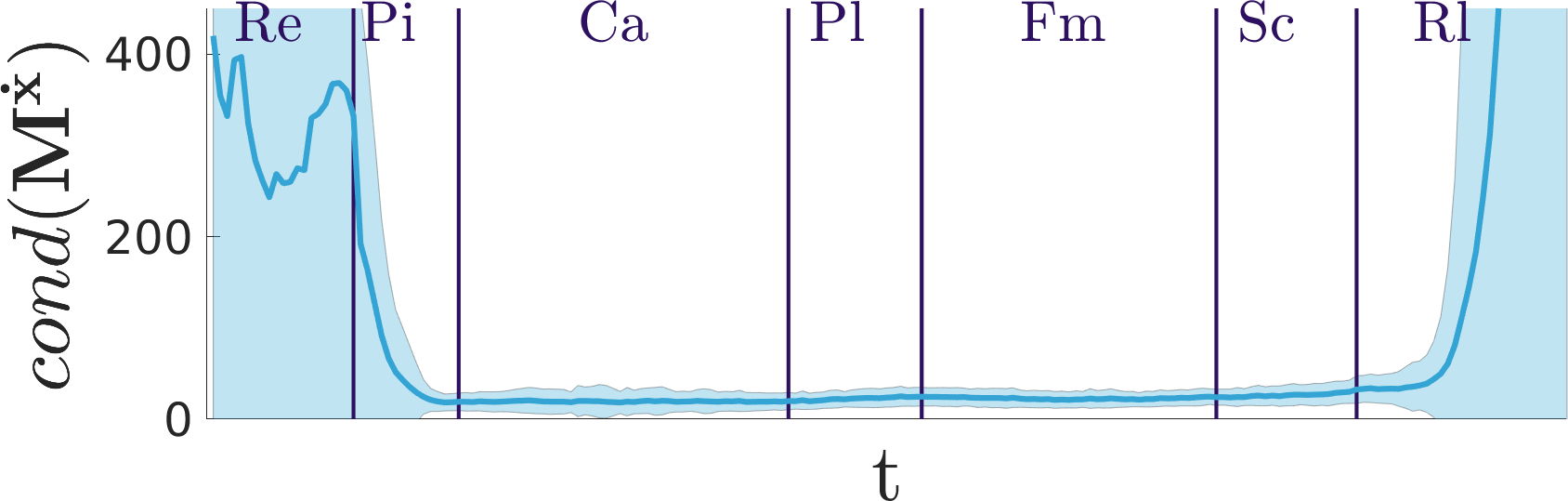}
		\caption{Screw middle $\mathsf{SM}$}
		\label{subFig:SMall}
	\end{subfigure}
	\begin{subfigure}[b]{0.32\textwidth}
		\includegraphics[width=\textwidth]{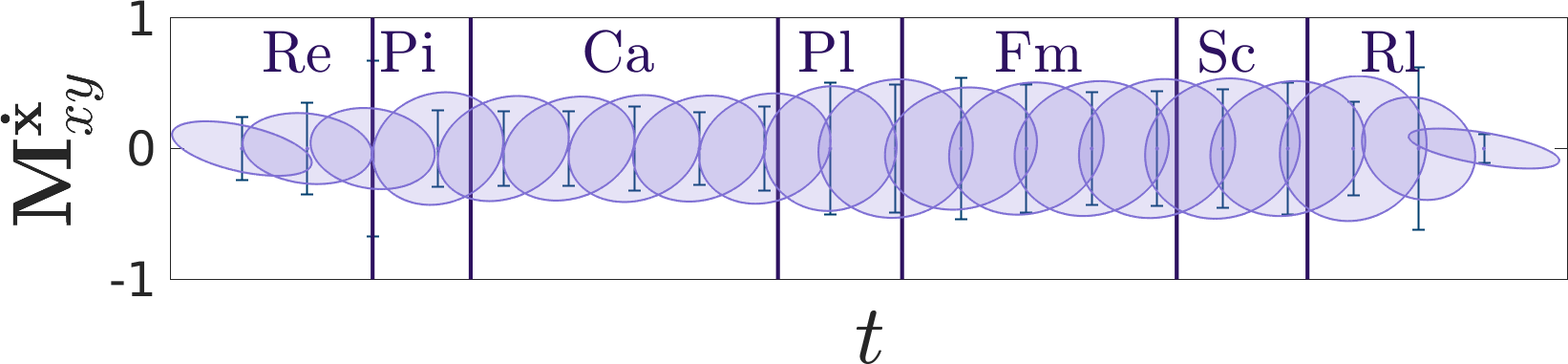}
		
		\vspace{0.1cm}
		\includegraphics[width=\textwidth]{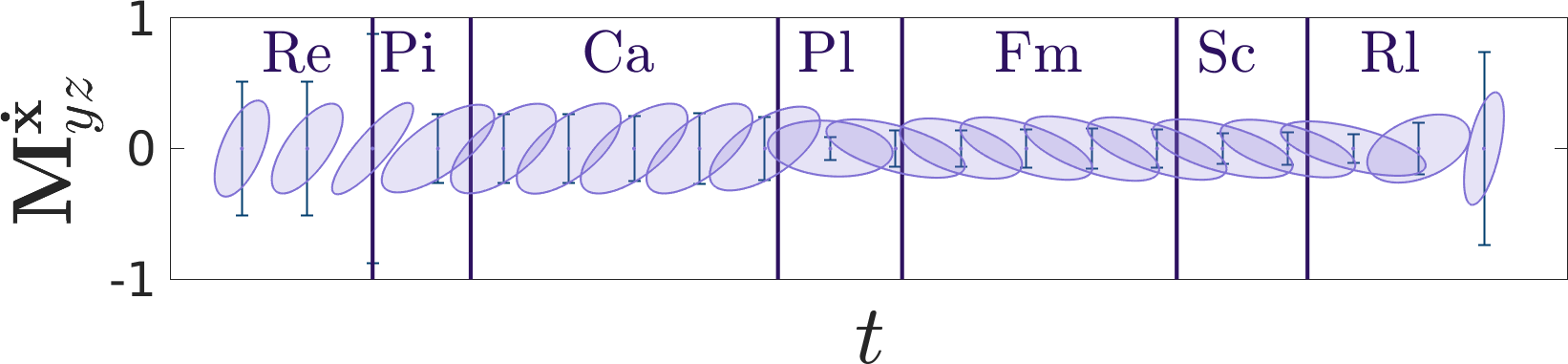}
		
		\vspace{0.1cm}
		\includegraphics[width=\textwidth]{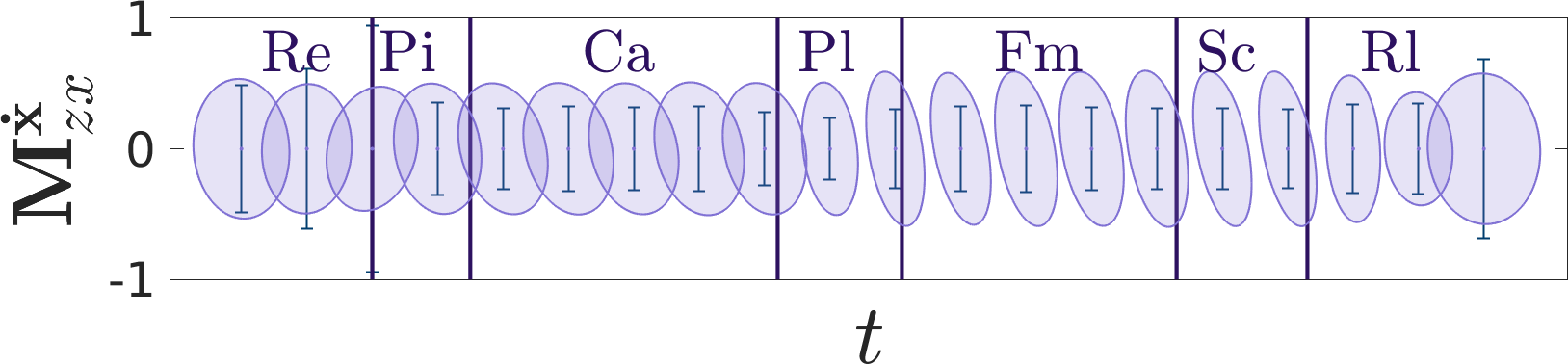}
		
		\vspace{0.1cm}
		\includegraphics[width=\textwidth]{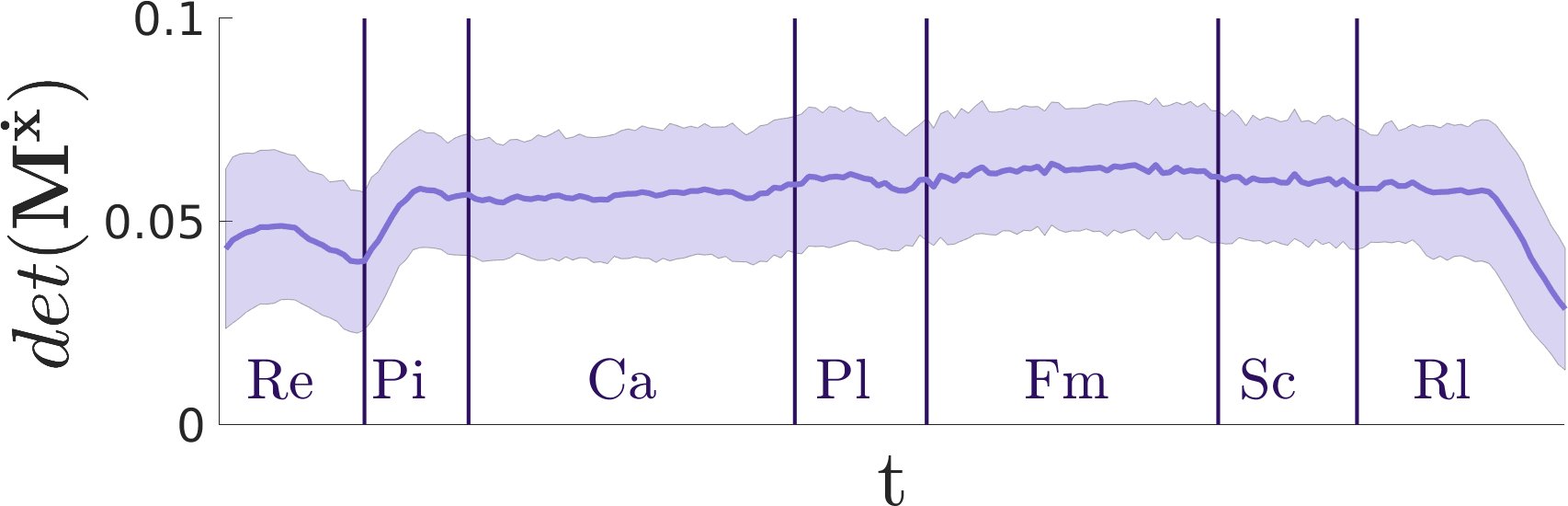}
		
		\vspace{0.1cm}
		\includegraphics[width=\textwidth]{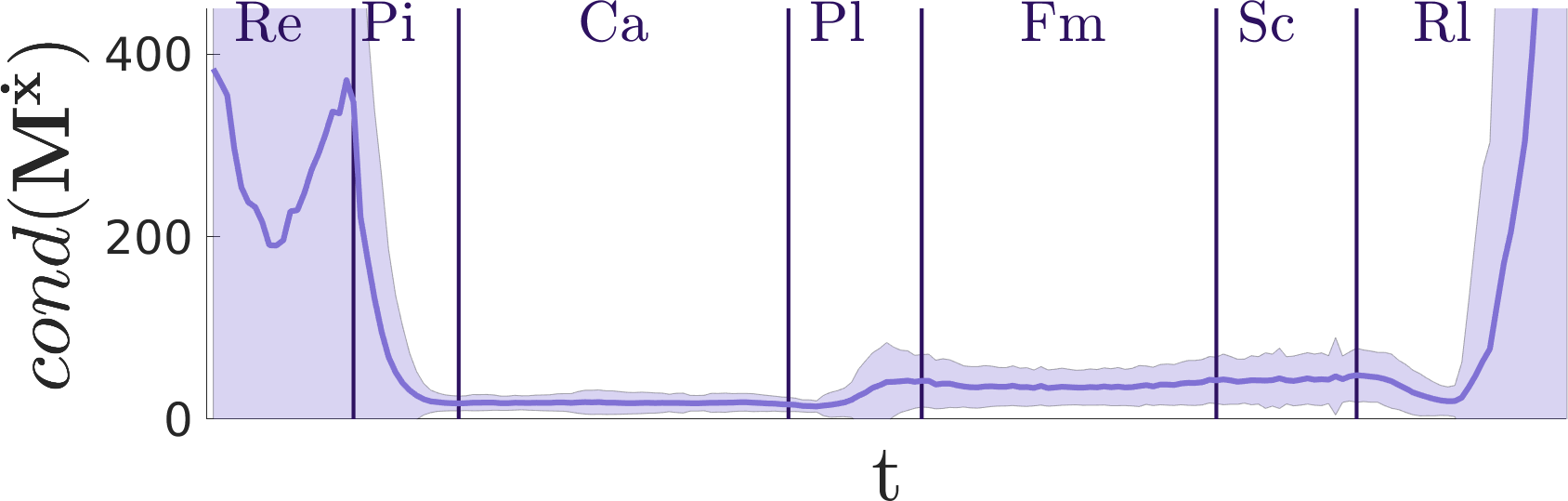}
		\caption{Screw high $\mathsf{SH}$}
		\label{subFig:SHall}
	\end{subfigure}
	\caption{Temporal evolution of the single-arm velocity manipulability ellipsoid for the screwing tasks. The inter-participants statistics are displayed with ellipsoids represented w.r.t the shoulder reference frame. The three first rows depict 2D-projections of the ellipsoids mean, along with 3 standard deviations of the vertical axis of the ellipsoid for the specific graph. The two bottom rows display the mean and standard deviation of the determinant and condition number of the ellipsoid, respectively.}
	\label{Fig:ScrewingTask}
	\vspace{-0.4cm}
\end{figure*}

To study the evolution of the manipulability during the screwing task, a subset of manipulability ellipsoids equally spaced in time is first selected by subsampling the manipulability sequence of each action of the task. This results in a dataset containing the same number of manipulability ellipsoids for each action across all trials. The evolution of the right-arm velocity manipulability ellipsoid during the different screwing tasks is studied in Fig.~\ref{Fig:ScrewingTask}. All the graphs show inter-participants statistics and the ellipsoids are represented w.r.t. the shoulder reference frame. The three $\emph{top}$ graphs of each column display the mean of the velocity manipulability ellipsoid profile for the corresponding task computed by~\eqref{Eq:SPDmean}. In each graph, the standard deviation of the ellipsoid along the vertical axis, equal to the square root of the corresponding diagonal element of the covariance tensor~\eqref{Eq:SPDcov}, is represented with error bars. For completeness, the two $\emph{bottom}$ graphs of each column show the mean and standard deviation of two classical manipulability indices, namely $\det(\bm{M})$ and $\text{cond}(\bm{M})$, denoting the determinant and condition number of $\bm{M}$, respectively. The former approximates the manipulability ellipsoid volume, while the latter relates to the ellipsoid isotropy.

Interestingly, we observe that the evolution of the manipulability mean during the task is consistent across the screwing motions. The beginning and end of the task, namely $\Rec$ and $\Rl$ actions, are characterized by narrow ellipsoids along the $y$ axis (i.e. the vertical direction), due to the fact that the arm rests along the body. These also correspond to actions displaying the highest variance. As in Fig.~\ref{Fig:HumanScrewing}, the velocity manipulability mainly narrows along $z$ and elongates along $x$ and $y$ during $\Fm$ and $\Sc$ actions for the three screwing heights. Therefore, the manipulability of the participants is generally adapted to the task. Moreover, we notice that the ellipsoid shape along $x$ and $y$ is generally similar from $\Ca$ to $\Sc$ actions. However, the manipulability ellipsoid is more isotropic for carrying actions $\Ca$. This is due to the fact that the participants usually prepare for screwing after having picked the screw and the bolt, i.e., they do not put back their arms at a neutral resting position but instead keep them in front of their torso. Therefore, these manipulability ellipsoids indicate an arm posture adaptation that anticipates the next action (i.e. planning phase), whose manipulability requirements are more specific.

Another relevant observation is that the classical manipulability indices, namely the determinant and condition number of the ellipsoids, tend to remain nearly constant during the whole movement, except in the reaching ($\Rec$) and releasing ($\Rl$) phases. In contrast, as emphasized previously, the shape of the velocity manipulability ellipsoid varies consistently during the different actions of the screwing task. Then, the determinant and condition number are uninformative measures that prohibit a proper human manipulability analysis.

Note that a similar analysis may be conducted for the velocity manipulability ellipsoids of the left arm for the screwing tasks. The left arm manipulability evolves similarly as the one of the right arm along the different actions. This is expected due to the presence of strong symmetries between the arms in this particular task. 

\begin{figure*}[tbp]
	\centering
	\begin{subfigure}[b]{0.16\textwidth}
		\includegraphics[width=\textwidth]{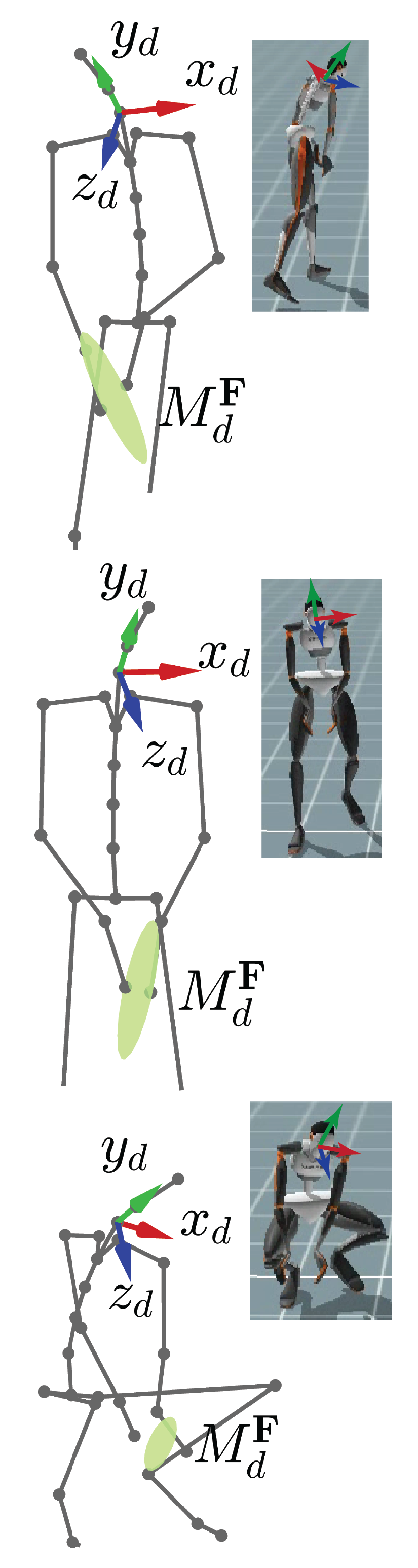}
		\caption{5kg ($\mathsf{C5}$)}
		\label{subFig:HumanC5}
	\end{subfigure}
	\begin{subfigure}[b]{0.32\textwidth}
		\includegraphics[width=\textwidth]{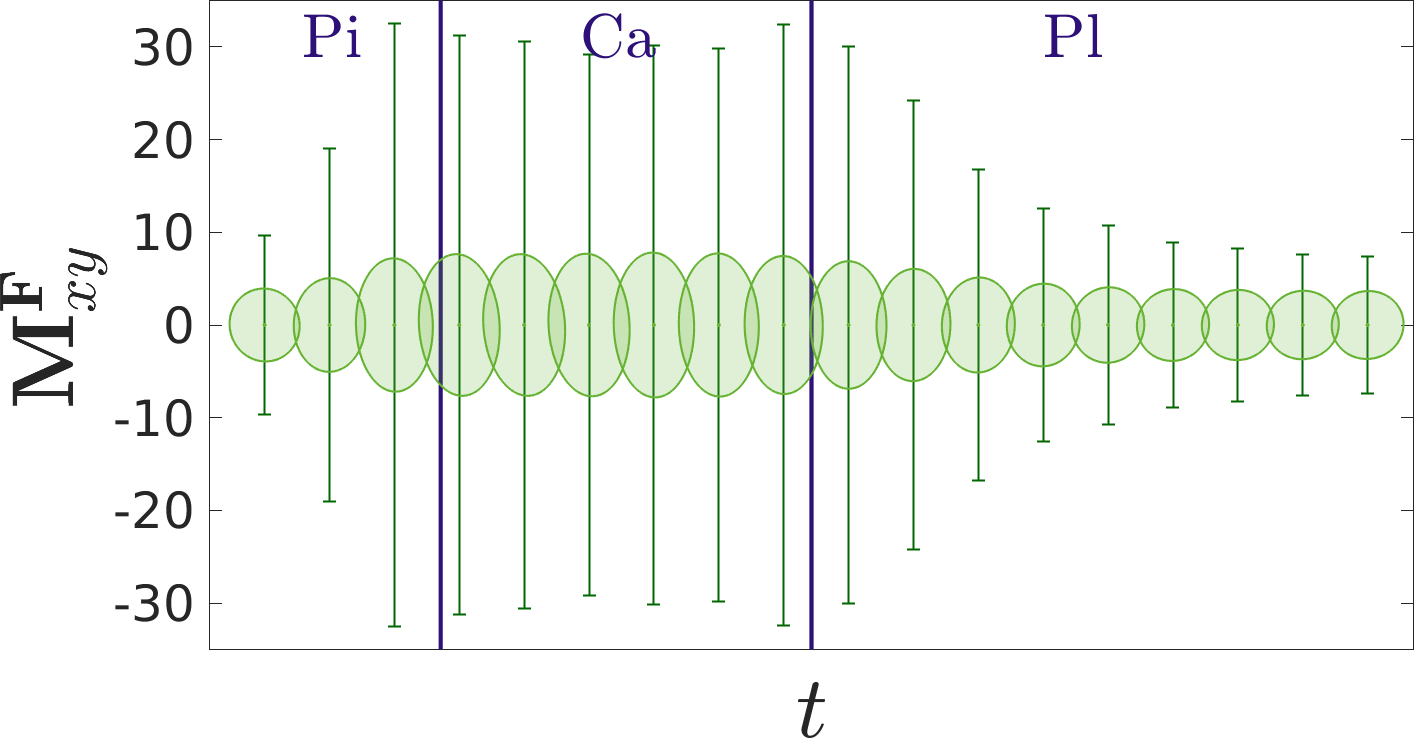}
		
		\vspace{0.1cm}
		\includegraphics[width=\textwidth]{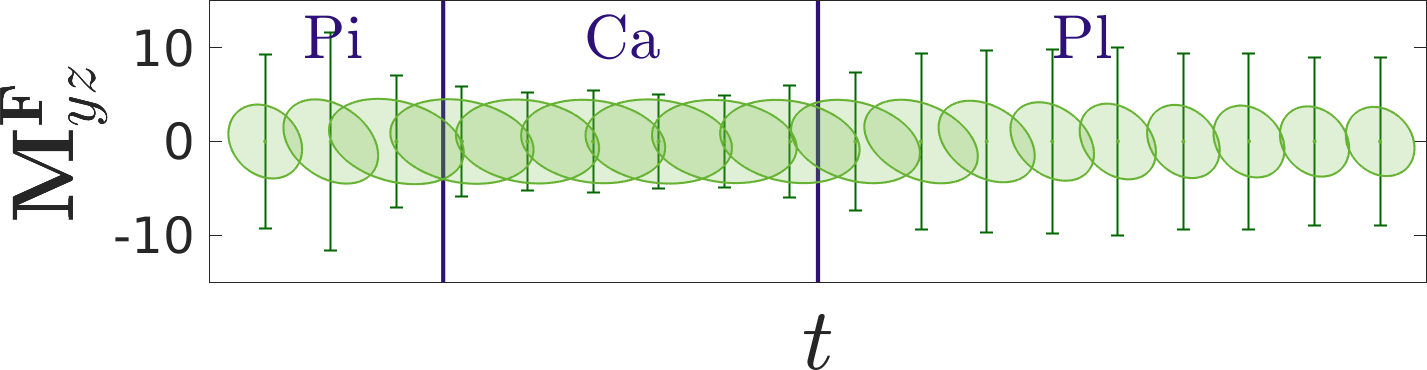}
		
		\vspace{0.1cm}	
		\includegraphics[width=\textwidth]{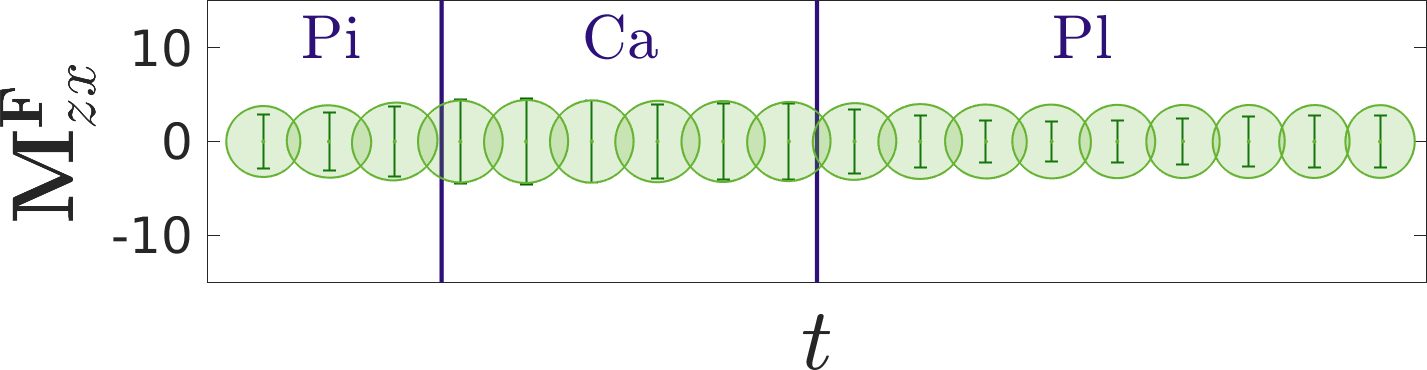}
		
		\vspace{0.1cm}
		\includegraphics[width=\textwidth]{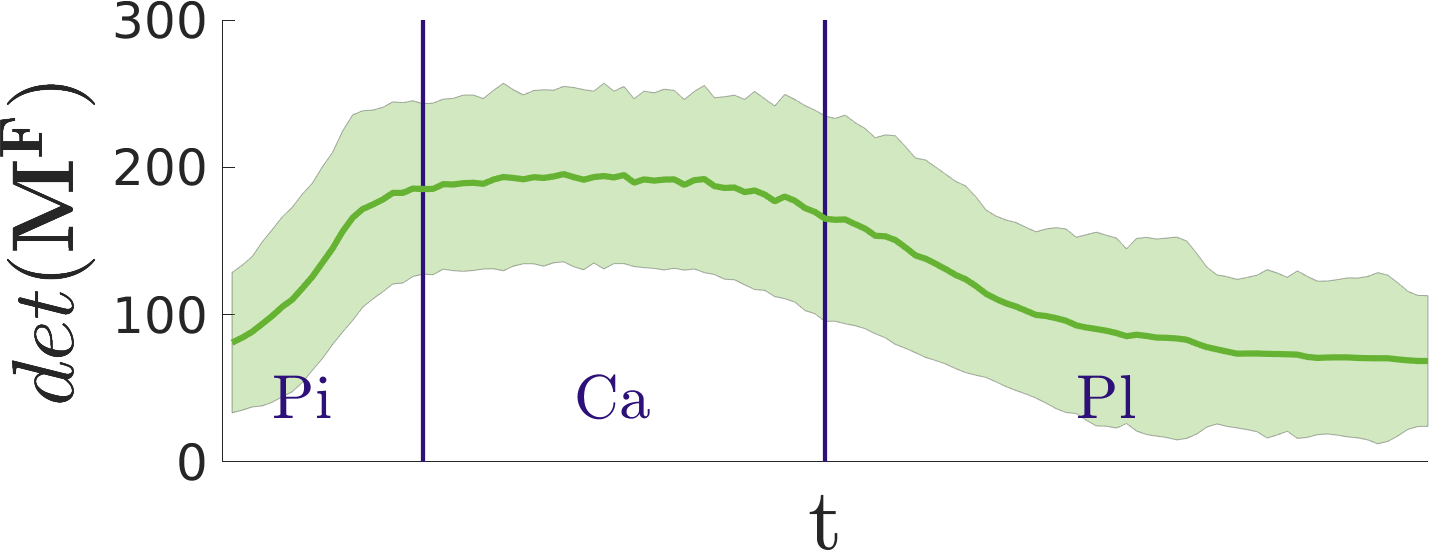}
		
		\vspace{0.1cm}
		\includegraphics[width=\textwidth]{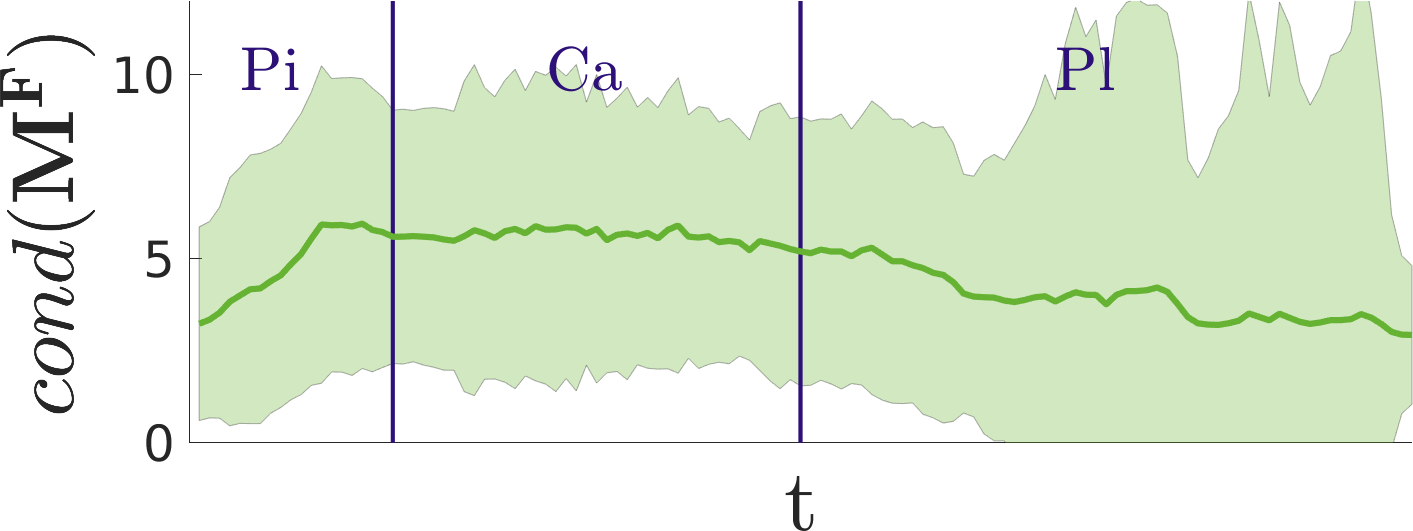}
		\caption{5kg ($\mathsf{C5}$)}
		\label{subFig:C5all}
	\end{subfigure}
	\begin{subfigure}[b]{0.32\textwidth}
		\includegraphics[width=\textwidth]{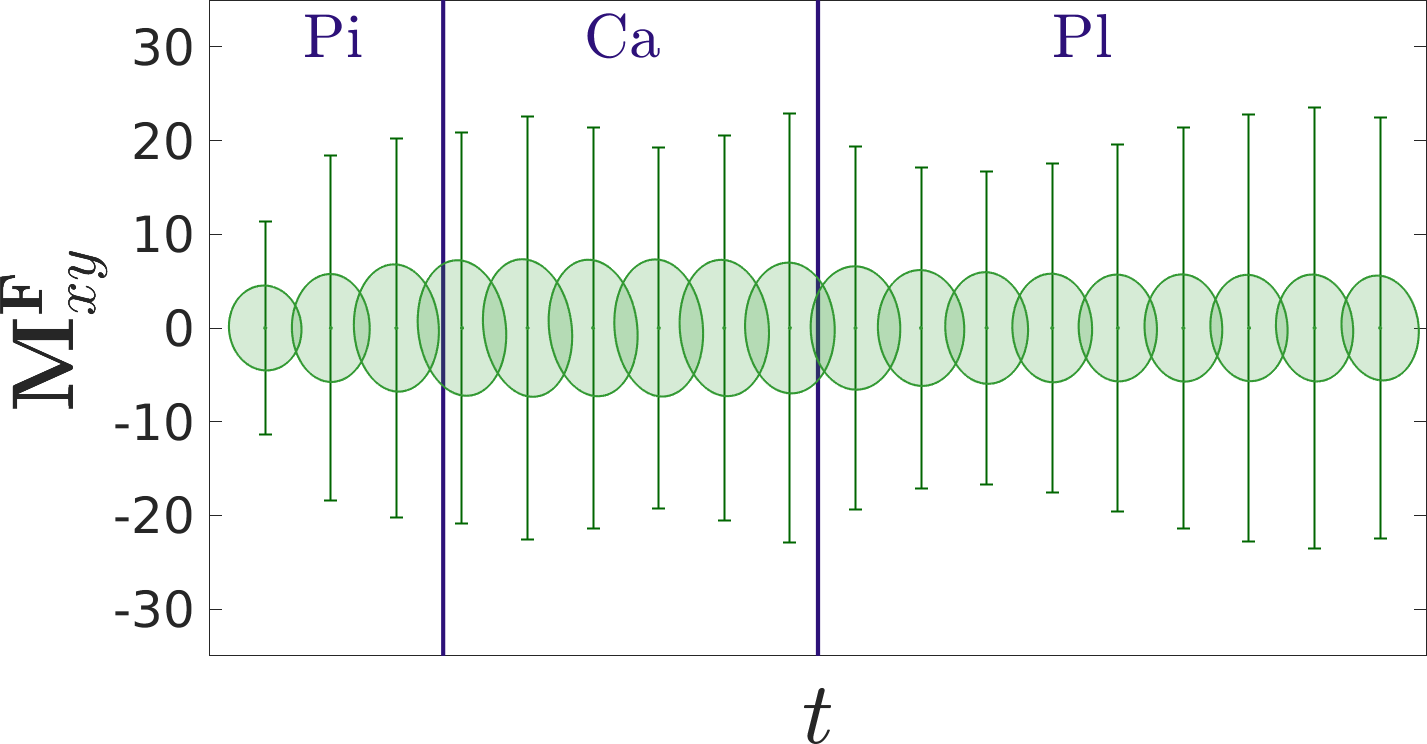}
		
		\vspace{0.1cm}
		\includegraphics[width=\textwidth]{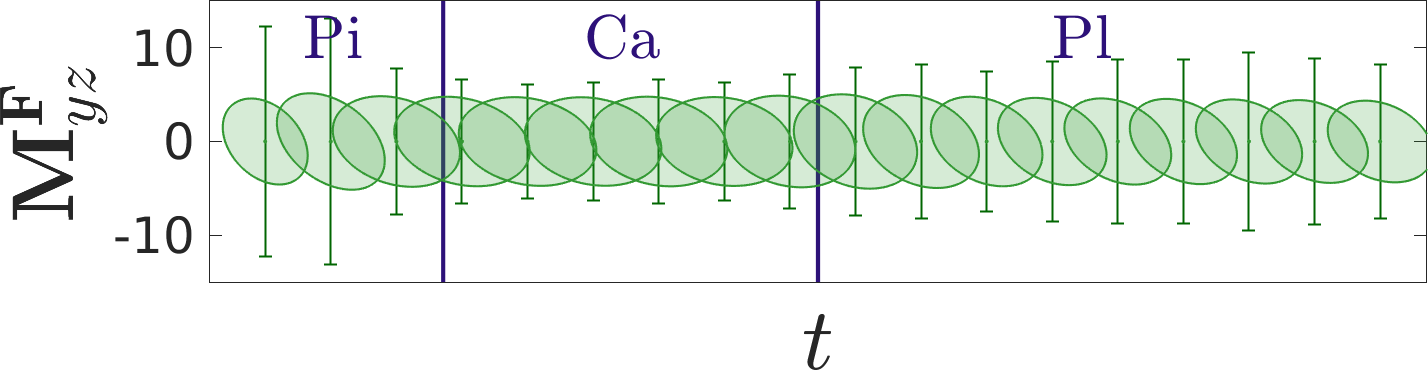}
		
		\vspace{0.1cm}
		\includegraphics[width=\textwidth]{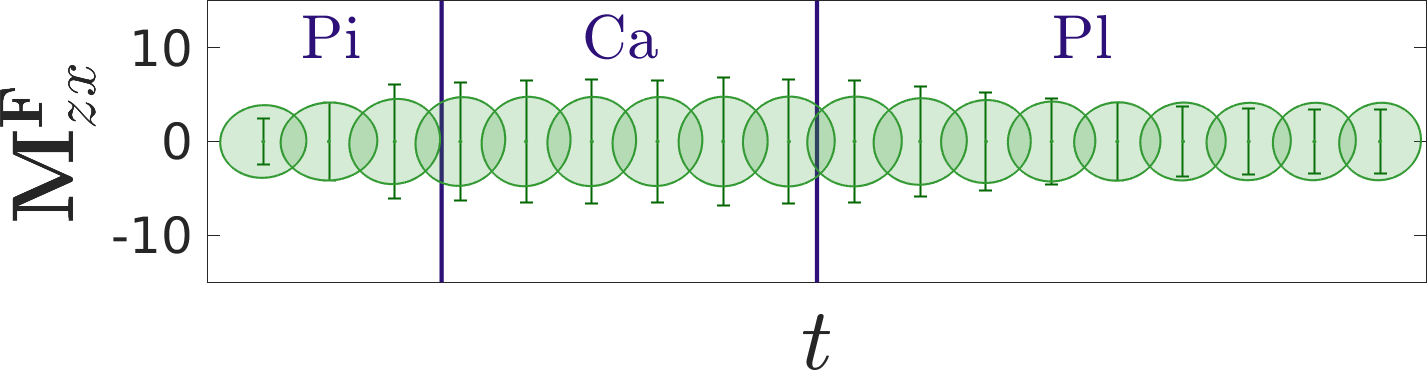}
		
		\vspace{0.1cm}
		\includegraphics[width=\textwidth]{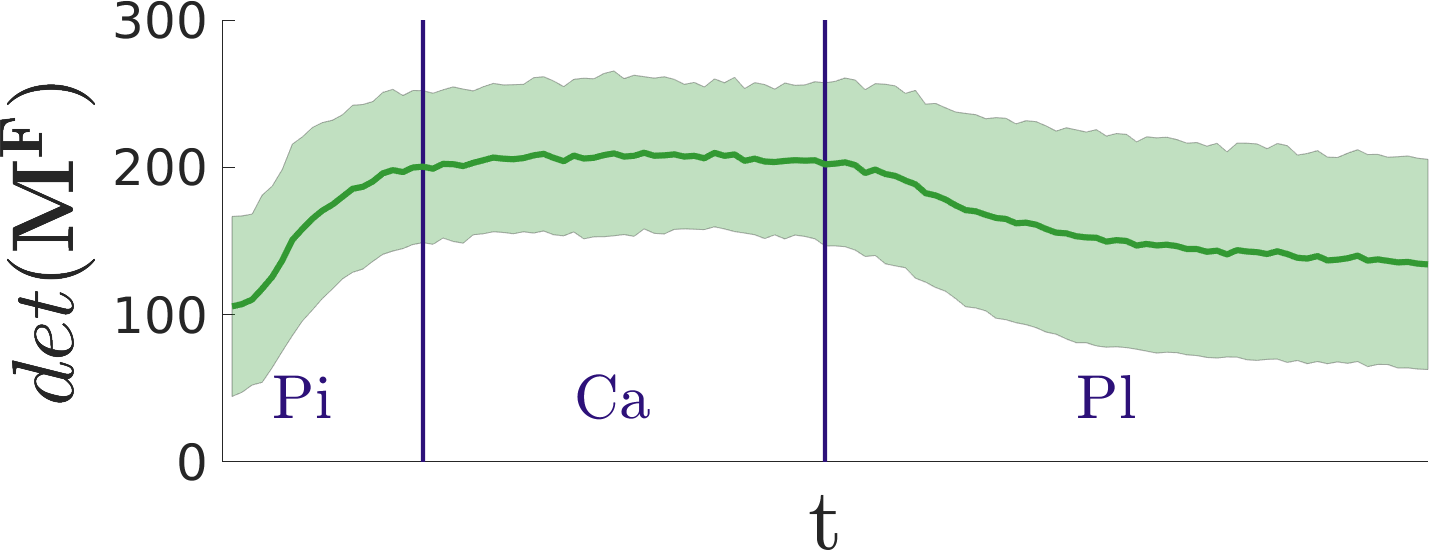}
		
		\vspace{0.1cm}
		\includegraphics[width=\textwidth]{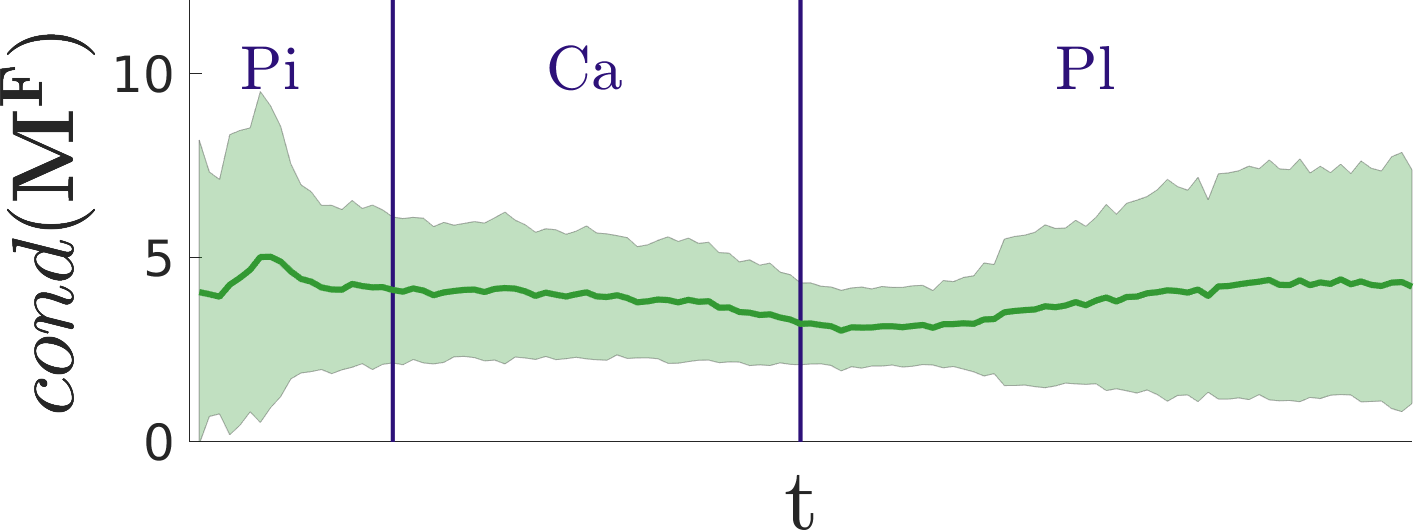}
		\caption{10kg ($\mathsf{C10}$)}
		\label{subFig:C10all}
	\end{subfigure}
	\begin{subfigure}[b]{0.16\textwidth}
		\includegraphics[width=\textwidth]{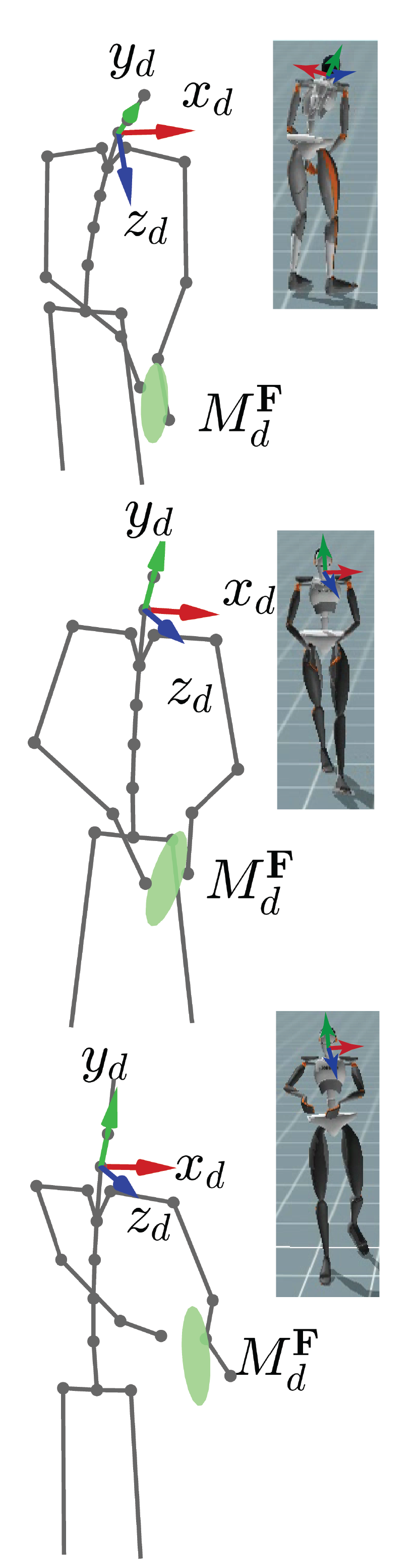}
		\caption{10kg ($\mathsf{C10}$)}
		\label{subFig:HumanC10}
	\end{subfigure}
	\caption{Temporal evolution of the dual-arm force manipulability ellipsoid for carrying tasks. \emph{(a)} and \emph{(d)} depict the posture of the participant 541 during the $\Pic$, $\Ca$ and $\Pl$ actions (from top to bottom). The dual-arm force manipulability ellipsoids (scaled by a factor 0.03) are depicted along with the neck reference frame. The three first rows of \emph{(b)} and \emph{(c)} show 2D projections of the ellipsoids mean, along with 1 standard deviation of the vertical axis of the ellipsoid for the specific graph. The two bottom rows of \emph{(b)} and \emph{(c)} display the mean and standard deviation of the determinant and condition number of the ellipsoid, respectively. The inter-participant statistics are displayed with ellipsoids represented in the neck reference frame.}
	\label{Fig:CarryingTask}
	\vspace{-0.4cm}
\end{figure*}

Regarding the carrying task, we analyze the dual-arm force manipulability. Figures~\ref{subFig:HumanC5} and~\ref{subFig:HumanC10} show the posture of a participant executing the $\mathsf{C5}$ and $\mathsf{C10}$ tasks, respectively. Similarly to the results reported in Fig.~\ref{Fig:ScrewingTask} for the screwing task, Fig.~\ref{subFig:C5all} and~\ref{subFig:C10all} depict the evolution of the dual-arm force manipulability ellipsoid during the two carrying tasks. The inter-participant statistics are displayed and the ellipsoids are represented w.r.t. the neck reference frame.

We observe that the main axis of the force manipulability ellipsoids are clearly aligned with the vertical axis during the $\Ca$ action. Therefore, the posture adopted by the participants favor high force exertion along the vertical axis, which is necessary for carrying heavy loads. However, we do not distinguish consistent differences in the magnitude of the manipulability between the \SI{5}{\kilogram} and \SI{10}{\kilogram} loads. Furthermore, the manipulability ellipsoids are almost isotropic at both the beginning of the $\Pic$ action and during the $\Pl$ actions. This especially accentuates during $\Pl$ actions of the $\mathsf{C5}$ task. This may be attributed to the fact that the shelf where the load is placed is close to the ground, which may require a different posture for this specific action. Overall, higher variations are observed during the carrying tasks compared to those of the screwing tasks. This may be associated with less-strict constraints in the carrying task when contrasted with the screwing motions. Also, the perception of the load may differ across participants, influencing the adopted postures.

\section{Manipulability Transfer}
\label{sec:transfer}
In general, the posture of the robot may have an impact on its performance while executing a given task. Typically, tracking a manipulability profile adapted to the task may allow the robot to avoid singularity, handle perturbations during task execution, anticipate potential collisions, optimize the execution time or minimize the energy consumption.
Here, we illustrate how the manipulability analysis of human movements can be exploited to transfer manipulability-based posture variation to robots executing similar tasks without the need of complex kinematic mappings. Namely, we propose to transfer the manipulability requirements of screwing and carrying tasks ($\mathsf{SM}$ and $\mathsf{C5}$) from a human to a Centauro robot~\cite{Baccelliere2017}. To do so, we exploit the manipulability transfer framework introduced in~\cite{Rozo17IROS:ManTransfer,Jaquier20}, that allows robots to learn and reproduce manipulability ellipsoids from human demonstrations. 
For both tasks, the demonstrations consist of the $15$ recorded trials of the participant 541, previously used for the manipulability analysis. A subset of actions is considered for each task. The simulated experiments were performed using Pyrobolearn~\cite{Delhaisse2019:Pyrobolearn}.

Previously, we showed that a manipulability ellipsoid profile of the transition actions, such as the carrying action $\Ca$ for the screwing task, can be seen as part of a planning process for the next action of interest (i.e. screwing $\Sc$). We exploit this idea for the manipulability transfer of the screwing task. We propose to learn and reproduce the manipulability profile of the actions preceding $\Sc$, namely $\Ca$, $\Pl$ and $\Fm$, aiming at reaching an appropriate posture to efficiently execute the main task. 
To do so, a desired time-driven manipulability profile $\bm{\hat{M}}_t$ is first learned from the demonstrations with a Gaussian mixture model (GMM) on the SPD manifold as in~\cite{Rozo17IROS:ManTransfer}. The robot is then required to track the desired manipulability profile while keeping balance and positioning its end-effector at a specific location, whose height is equal to the one of the $\mathsf{SM}$ motion. We assume a task hierarchy that prioritizes the position control of the center of mass (CoM) over the support polygon and zero velocity at all contact points with the floor, while both the end-effector position and manipulability tracking are considered secondary tasks. 
The corresponding full joint velocity controller for legged robots is defined as (see~\cite{Mistry08} for details)
\begin{equation}
\bm{\dot{q}} = \left[\begin{matrix}
\bm{I}_{n\times n} \\
\bm{0}_{6\times n} \\
\end{matrix}\right] ^\trsp\,\left( \bm{J}_b^{\dagger} \bm{\dot{x}}_b + \bm{N}_b \bm{\dot{q}}_{\bm{N}_b} \right),
\label{Eq:CoMtracking}
\end{equation} 
where the first term accounts for the virtual joints of a floating-base robot, $n$ is the number of DoF of the robot, $\bm{\dot{x}}_b = \left( \bm{0}^\trsp,
\bm{\dot{x}}_{\text{CoM}}^\trsp \right)^\trsp$ with $\bm{\dot{x}}_{\text{CoM}}$ the velocity at the robot CoM, 
$\bm{J}_b = \left( \bm{J}_{\text{feet}}^\trsp, \bm{J}_{\text{CoM,xy}}^\trsp \right)^\trsp$ is the Jacobian of the balancing task, $\bm{N}_b$ is the corresponding nullspace projection matrix and $\bm{\dot{q}}_{\bm{N}_b}$ is the joint velocities of the secondary tasks.
In the first part of the motion, corresponding to the carrying action $\Ca$, the manipulability tracking is prioritized over the position tracking in order to allow the robot to reach a good initial posture for executing the task. Therefore, we defined the joint velocities of the secondary task as
\begin{multline}
\bm{\dot{q}}_{\bm{N}_b}\!=\!(\bm{\mathcal{J}}_{(3)}^\dagger)^\trsp \, \bm{K}_{\bm{M}} \, \text{vec}\Big(\!\text{Log}_{\bm{M}}(\bm{\hat{M}})\!\Big) \\
+ \Big(\! \bm{I}- (\bm{\mathcal{J}}_{(3)}^\dagger)^\trsp \bm{\mathcal{J}}_{(3)}^\trsp\! \Big) \bm{J}^\dagger \, \bm{K}_{\bm{x}} \, (\bm{\hat{x}}-\bm{x}),
\label{Eq:ManipTrackNullspace}
\end{multline} 
where $\bm{\hat{x}}$ and $\bm{x}$ are the desired and current end-effector positions, $\bm{\hat{M}}$ and $\bm{M}$ are the desired and current end-effector manipulability, $\bm{J}$ is the end-effector Jacobian and $\bm{\mathcal{J}}_{(3)}$ is the end-effector manipulability Jacobian in matrix form. The matrices $\bm{K}_{\bm{M}}$ and $\bm{K}_{\bm{x}}$ are proportional gains for manipulability and position tracking control, respectively. 
Then, for the remaining part of the motion, the position tracking is prioritized so that the robot can reach the screwing location with a posture adapted to the task requirements. Thus, the corresponding joint velocities are given by
\begin{multline}
\bm{\dot{q}}_{\bm{N}_b} = \bm{J}^\dagger \, \bm{K}_{\bm{x}} \, (\bm{\hat{x}}-\bm{x}) \\
+ (\bm{I}-\bm{J}^\dagger\bm{J}) \, (\bm{\mathcal{J}}_{(3)}^\dagger)^\trsp \, \bm{K}_{\bm{M}} \, \text{vec}\Big(\text{Log}_{\bm{M}}(\bm{\hat{M}})\Big).
\label{Eq:ManipTrackSecTask}
\end{multline}
The controllers details can be found in~\cite{Jaquier20}.

\begin{figure*}[tbp]
	\centering
	\begin{subfigure}[b]{0.47\textwidth}
		\begin{minipage}{0.3\textwidth}
			\includegraphics[width=\textwidth]{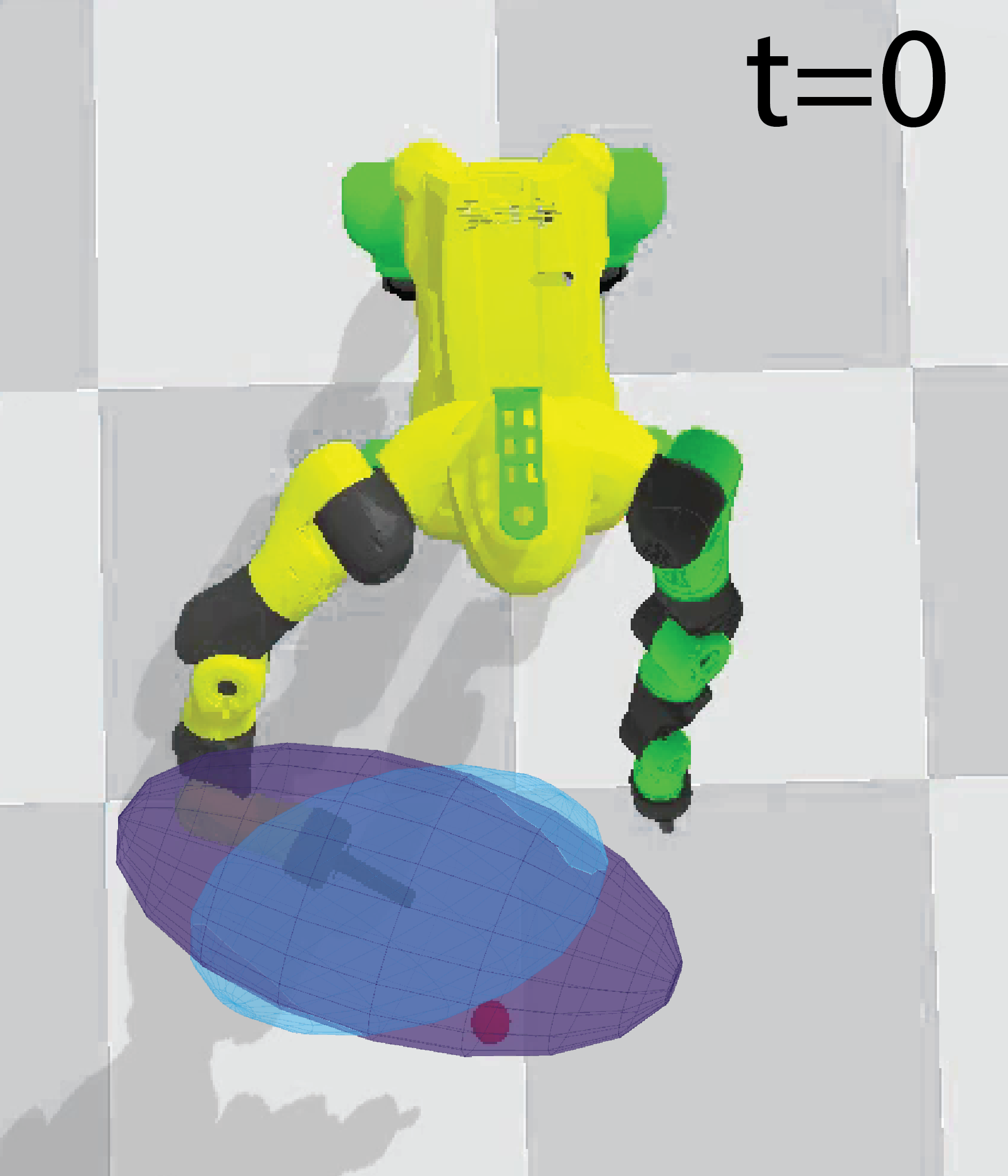}
			\includegraphics[width=\textwidth]{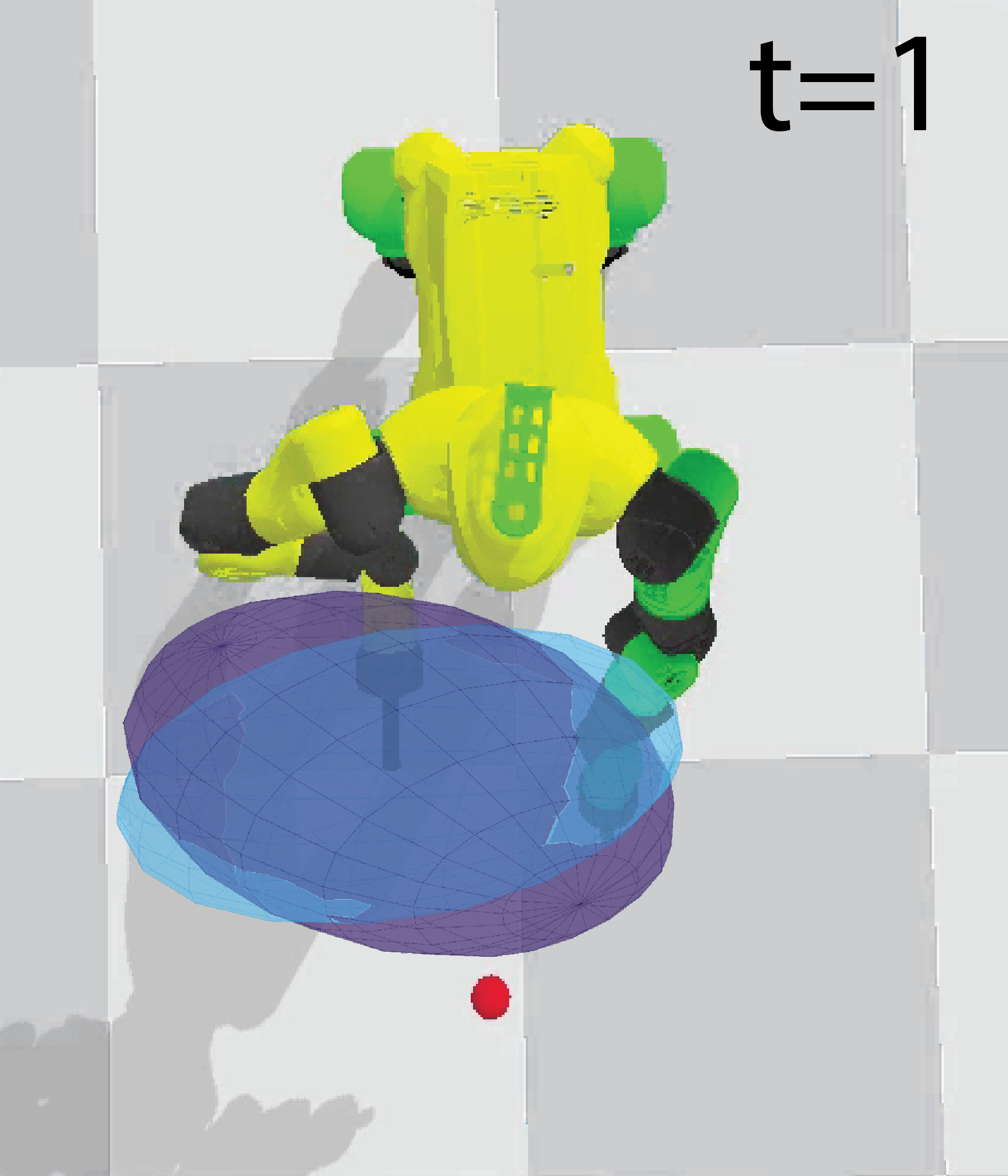}
			\includegraphics[width=\textwidth]{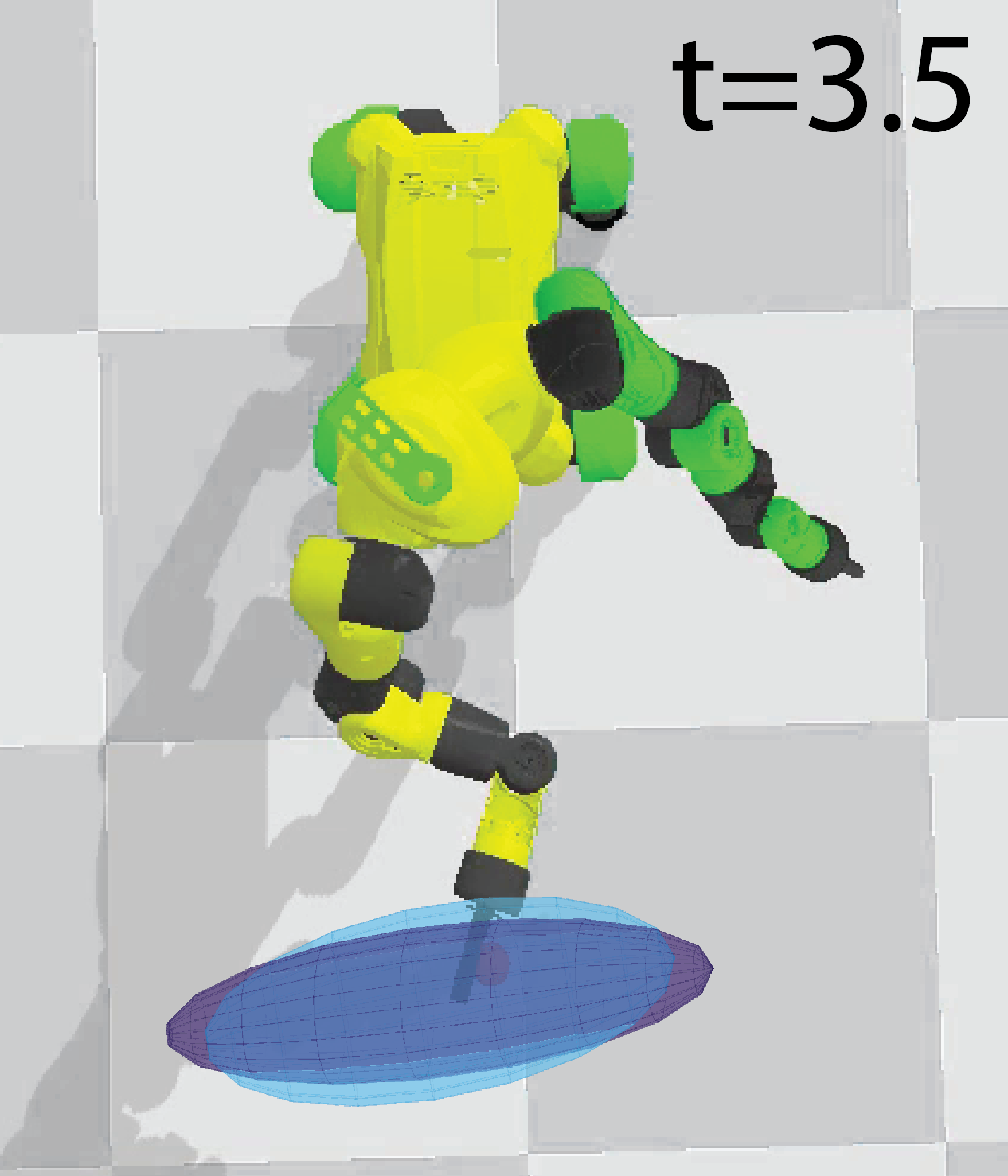}
		\end{minipage}
		\begin{minipage}{0.68\textwidth}
			\includegraphics[width=\textwidth]{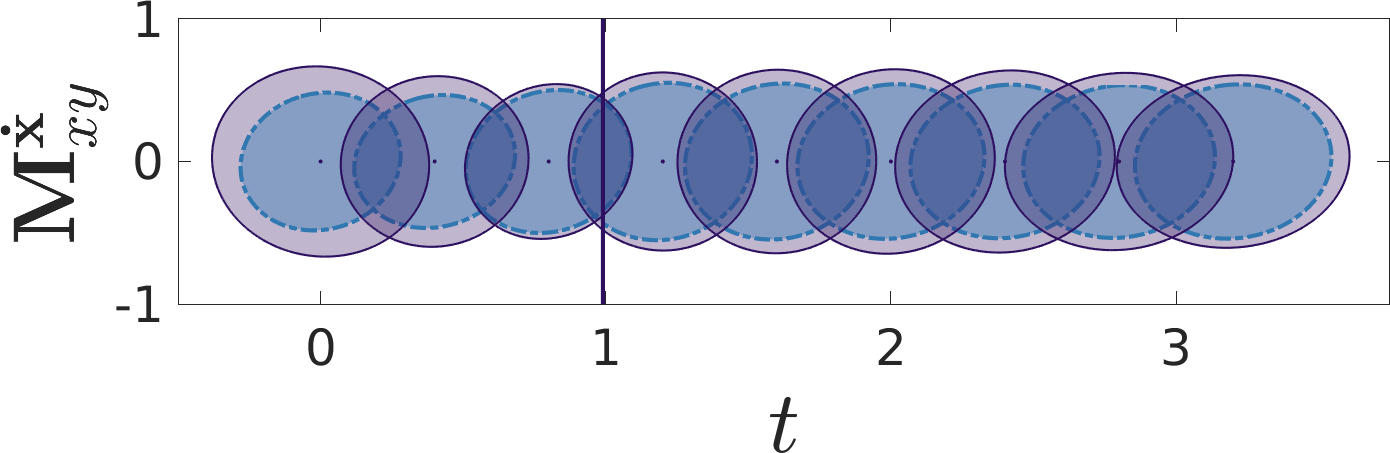}
			
			\vspace{0.1cm}
			\includegraphics[width=\textwidth]{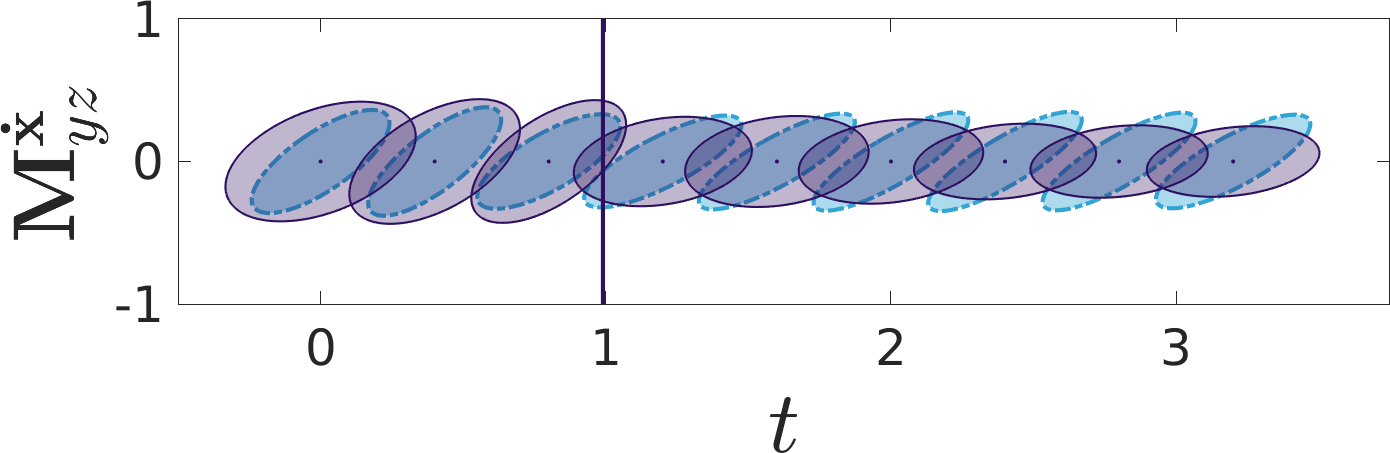}
			
			\vspace{0.1cm}
			\includegraphics[width=\textwidth]{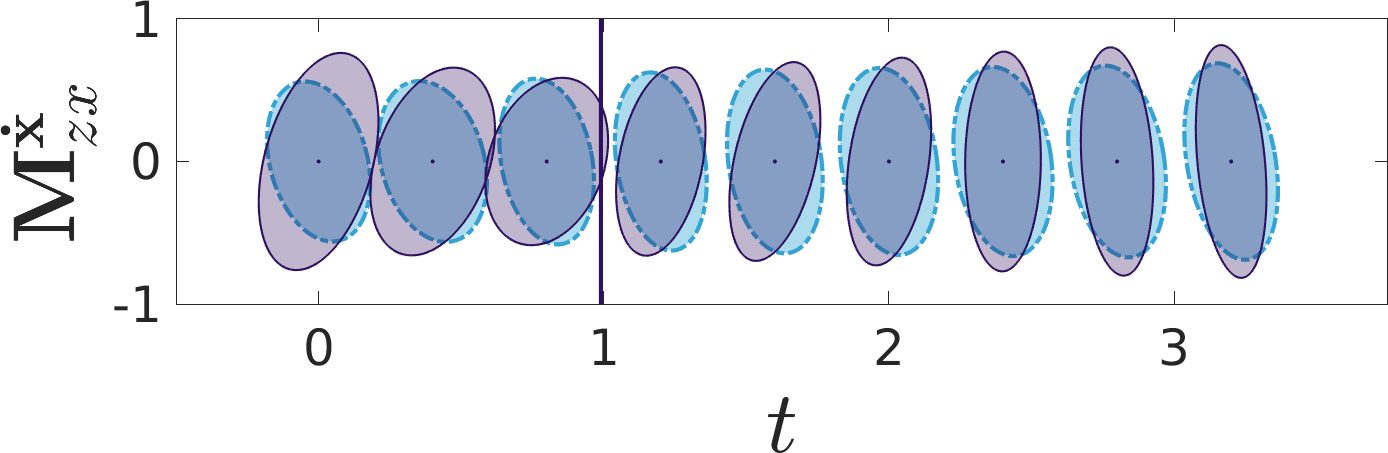}
			
			\vspace{0.1cm}
			\includegraphics[width=\textwidth]{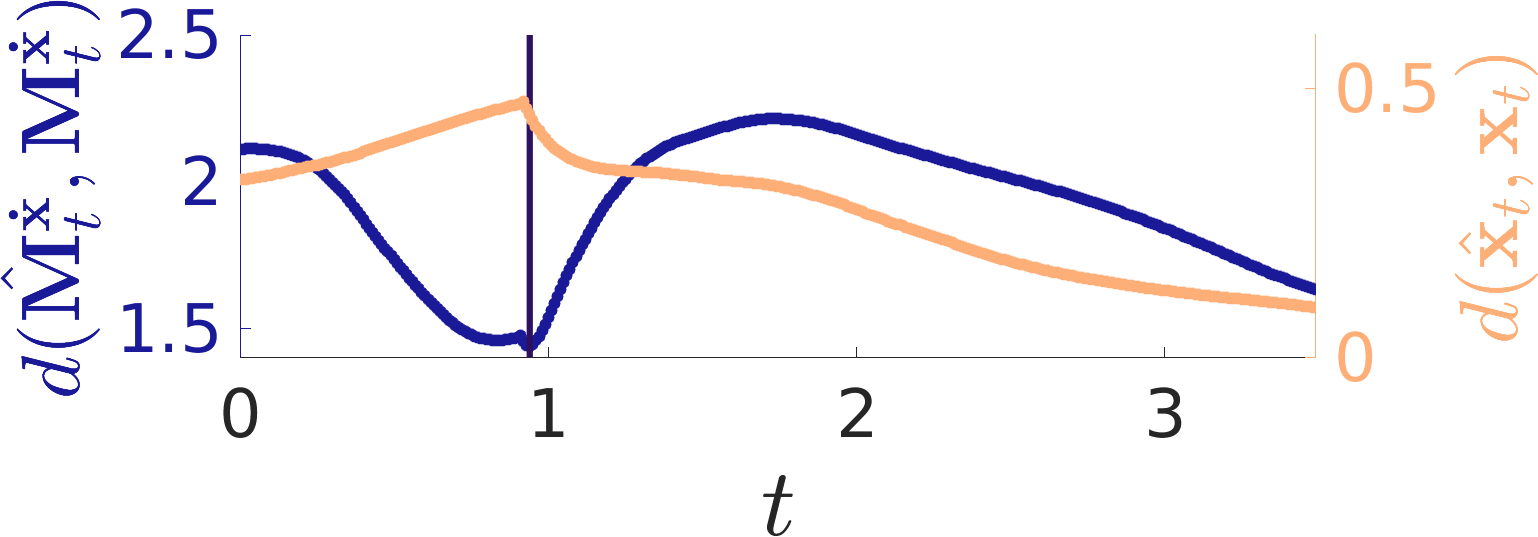}
		\end{minipage}
		\caption{$\mathsf{SM}$}
		\label{subFig:ManipTransferScrewing}
	\end{subfigure}
	\begin{subfigure}[b]{0.47\textwidth}
		\begin{minipage}{0.3\textwidth}
			\includegraphics[width=\textwidth]{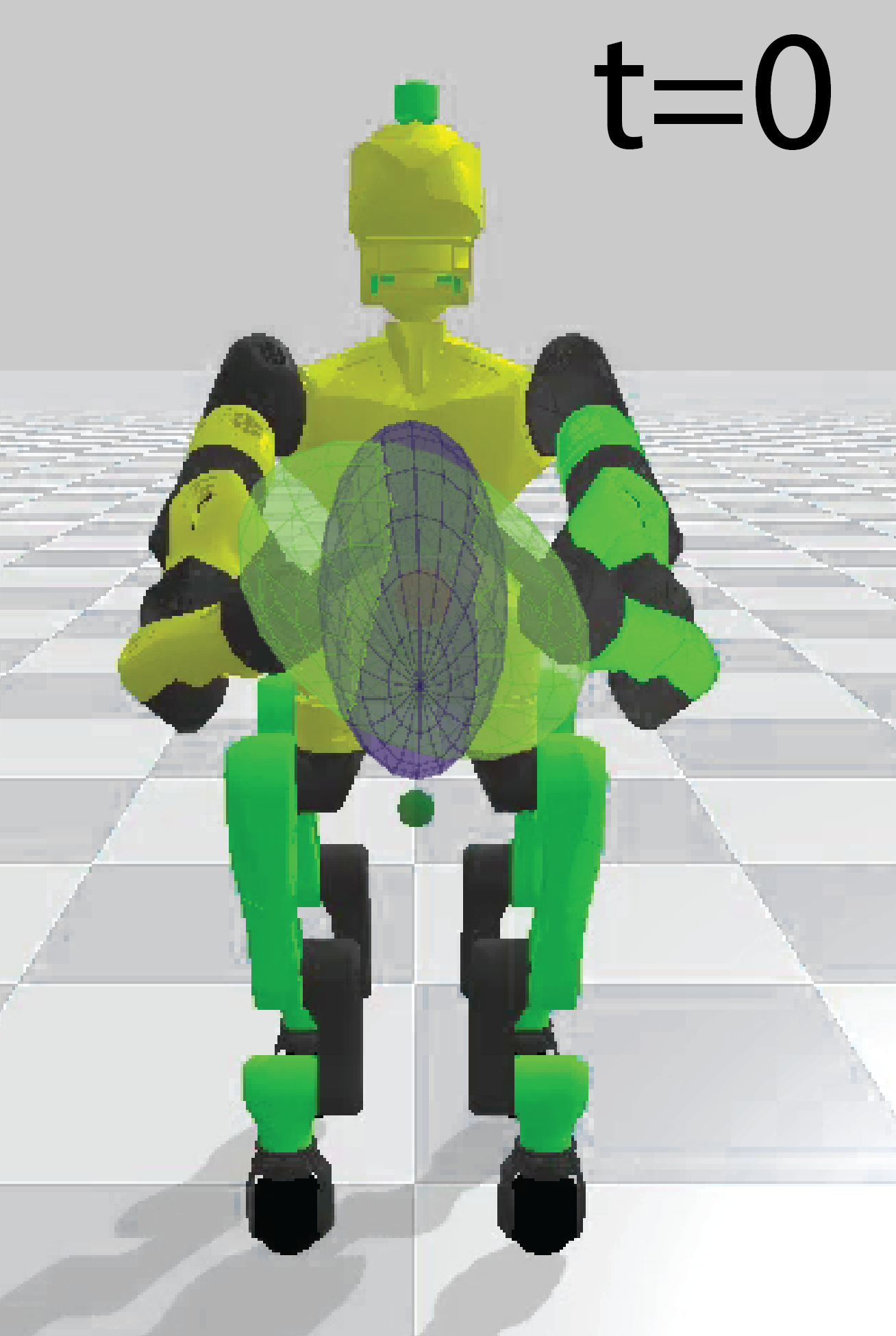}

			\vspace{0.1cm}
			\includegraphics[width=\textwidth]{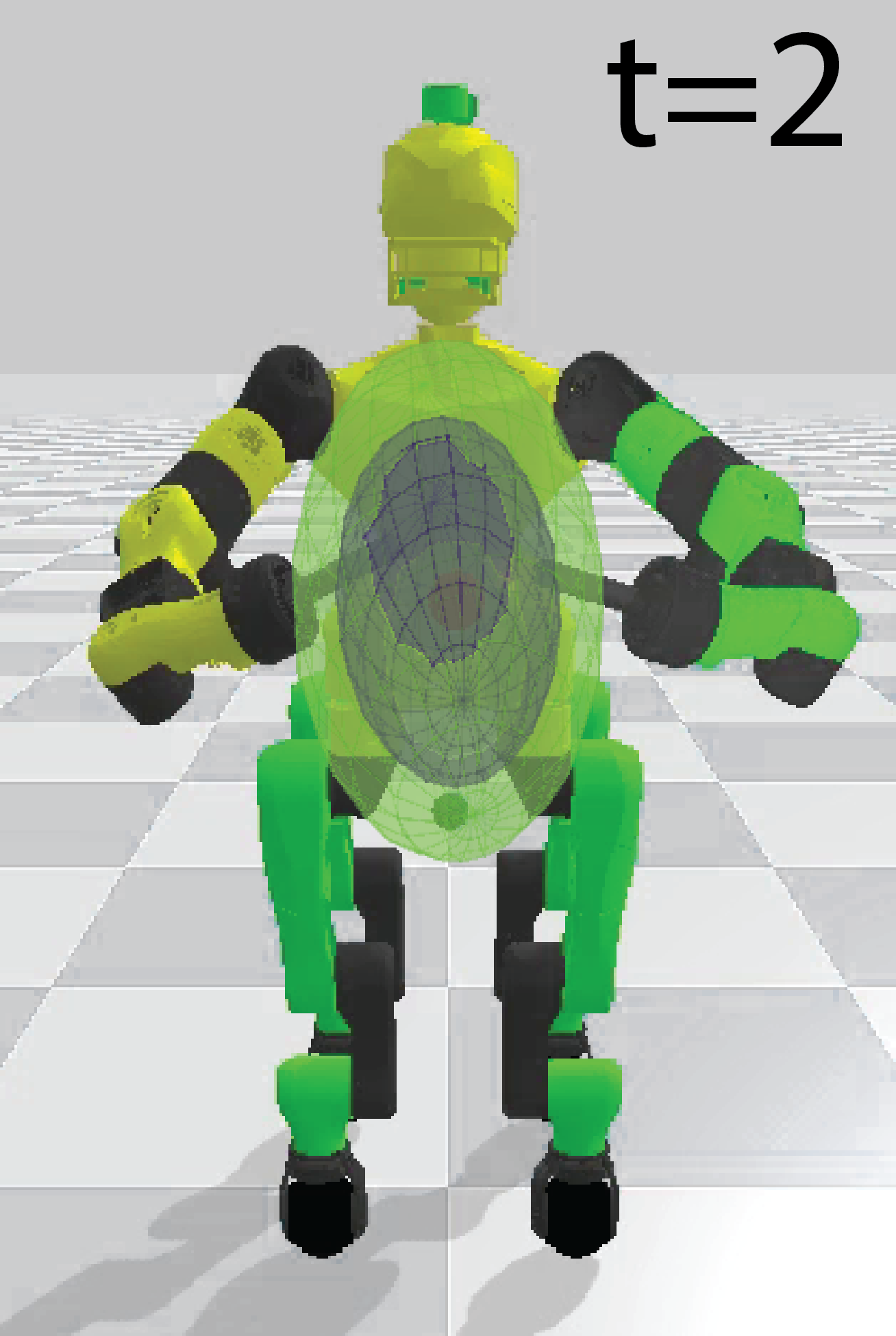}
		\end{minipage}
		\begin{minipage}{0.65\textwidth}
			\includegraphics[width=\textwidth]{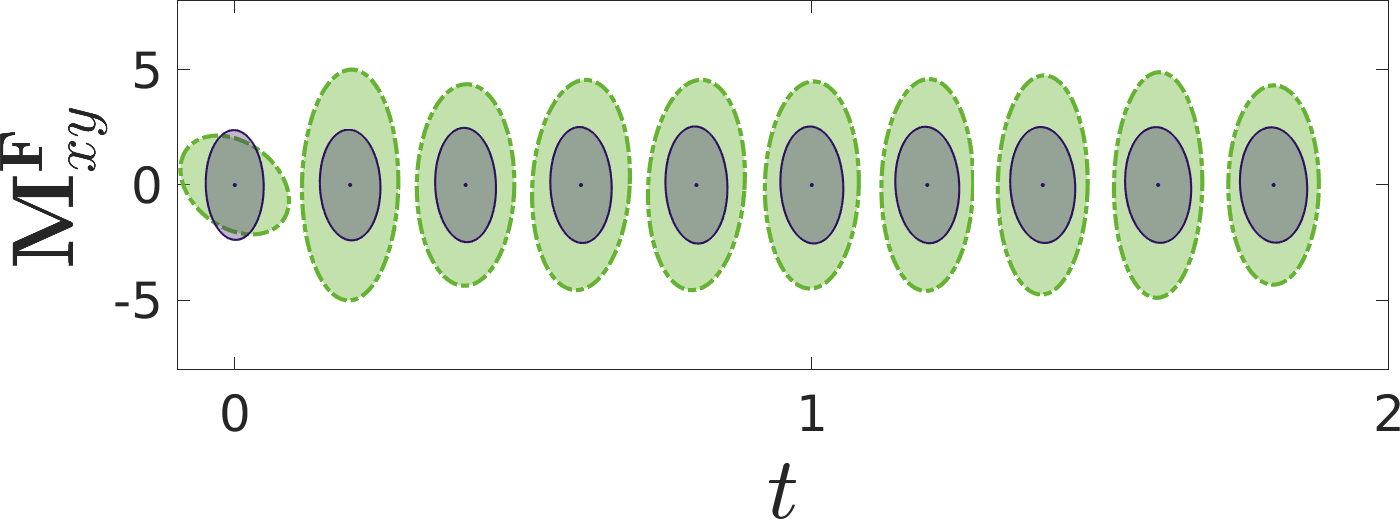}
			
			\vspace{0.1cm}
			\includegraphics[width=\textwidth]{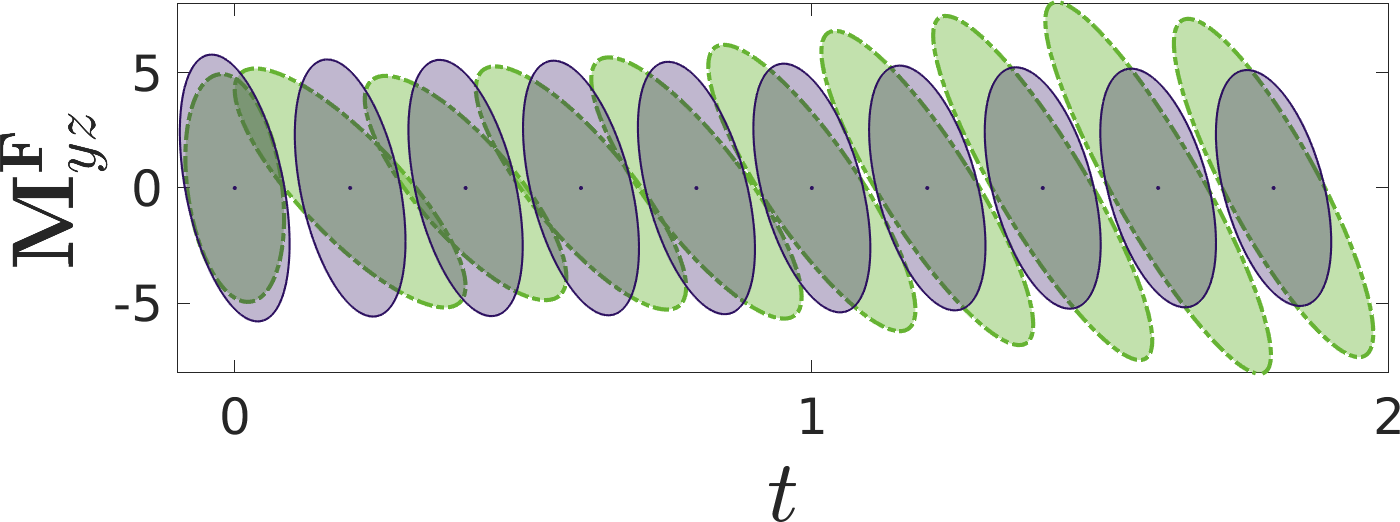}
			
			\vspace{0.1cm}
			\includegraphics[width=\textwidth]{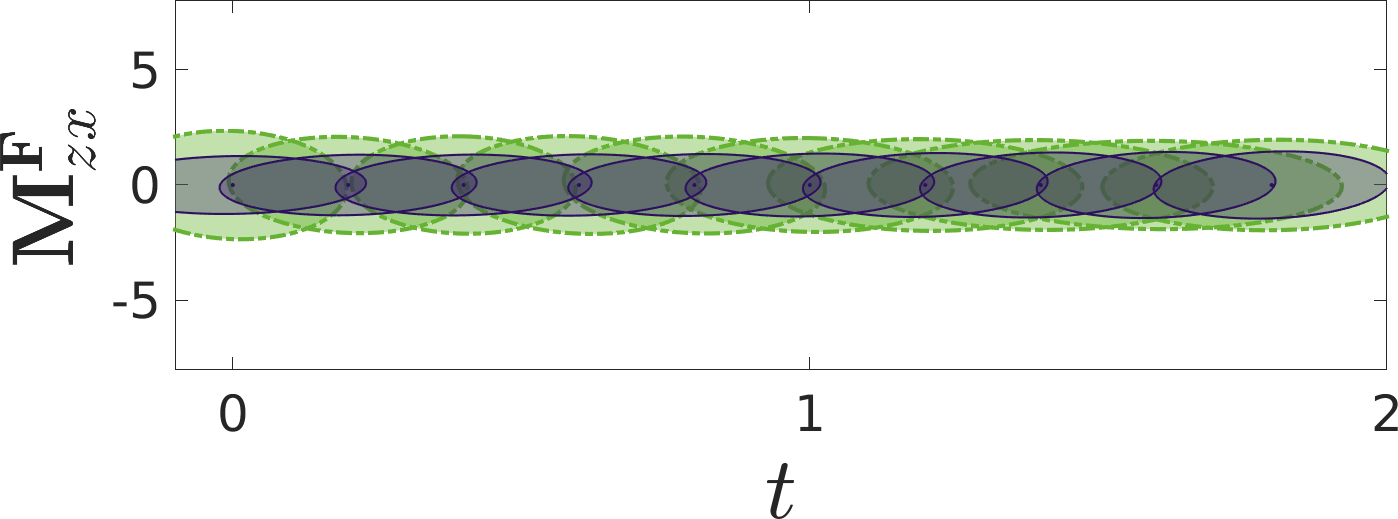}
			\includegraphics[width=\textwidth]{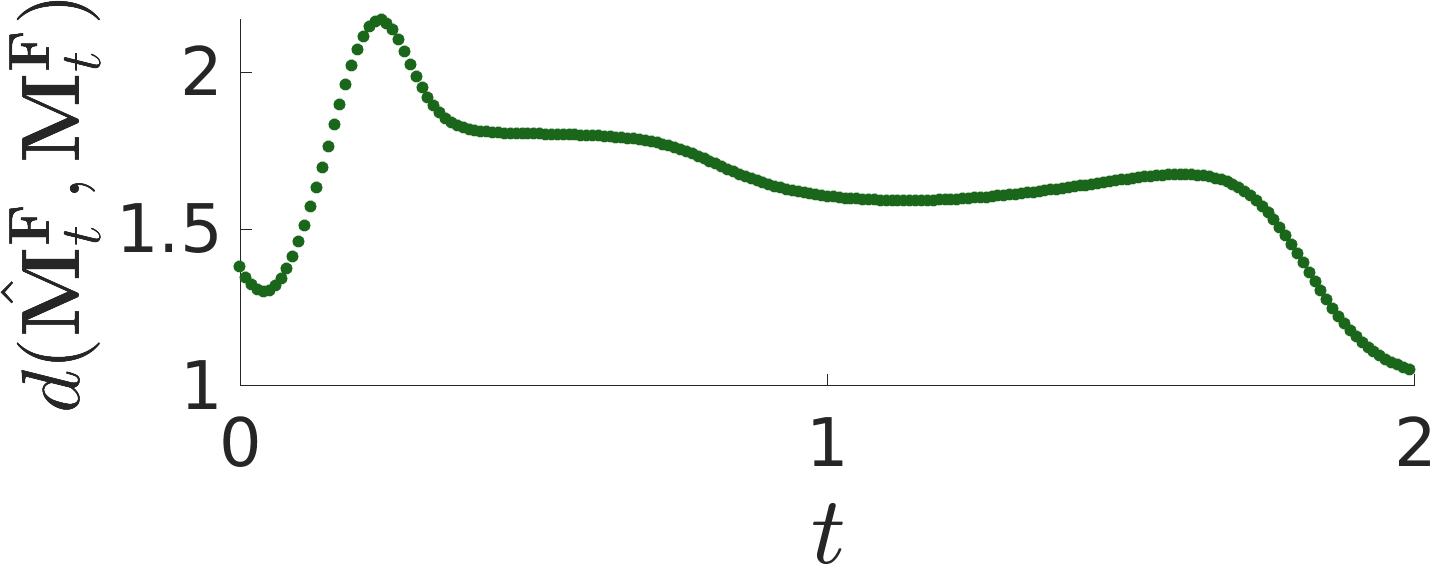}
		\end{minipage}
		\caption{$\mathsf{C5}$}
		\label{subFig:ManipTransferCarrying}
	\end{subfigure}
	\caption{Partial reproduction of industry-like tasks using human-robot manipulability transfer. Snapshots of the robot executing the tasks are shown on the \emph{left} part of the figures.
	The \emph{top-right} graphs shows the 2D projections of the time-varying desired manipulability profile, learned by demonstrations, and the reproduced manipulability depicted by \emph{(a)} blue, \emph{(b)} green and purple ellipsoids, respectively. The \emph{bottom-right} graph shows the distance between the current and desired manipulability (and end-effector position) over time (given in seconds). The change of controller during the pre-screwing motion is indicated by a vertical line.}
	\label{Fig:ManipTransfer}
	\vspace{-0.4cm}
\end{figure*}

Figure~\ref{subFig:ManipTransferScrewing} shows the evolution of the posture and the manipulability of Centauro during the reproduction of the pre-screwing motion with its right arm. The manipulability tracking has priority over the position control by~\eqref{Eq:ManipTrackNullspace} from $t=0$ to $t=1$s, and the priority order is reversed for the remaining time using~\eqref{Eq:ManipTrackSecTask}. Therefore, during the first part of the motion, we can observe that the robot mainly adapts its posture to fulfill the manipulability requirements. This naturally results in the robot orienting its end-effector outwards w.r.t. its torso. In this phase, the position error slightly increases, while the manipulability error decreases. The robot starts the second part of the task by moving its right arm to decrease the position error. Interestingly, despite the desired position could be reached solely by extending the arm, the robot instead uses its torso to rotate the arm in order to reach the desired end-effector position while still tracking the desired manipulability accurately. Note that tracking the desired manipulability for this task naturally favors arm postures where the end-effector is oriented outwards, matching a screwing motion whose main direction is orthogonal to a vertical plane in the robot workspace. This corresponds to the recorded screwing motion of the participant 541. However, considering a precise orientation remains necessary for successfully executing a complete screwing task, which can be straightforwardly included in our control formulation.

For the transfer of the $\mathsf{C5}$ task, we consider the part of the motion where the human picks up the load to carry it, i.e. $\Pic$ and $\Ca$ actions. After having learned the desired dual-arm manipulability profile from demonstrations, we employ the main controller~\eqref{Eq:CoMtracking} with the nullspace controller~\eqref{Eq:ManipTrackSecTask} for the whole task while keeping the end-effectors position fixed. Note that we consider all the DoFs of the robot for computing the dual-arm manipulability ellipsoid, so that the dual-arm Jacobian is defined as $\bm{J}_d = (\bm{J}_l^\trsp, \bm{J}_r^\trsp)^\trsp$. Figure~\ref{subFig:ManipTransferCarrying} shows the evolution of the posture and the manipulability of Centauro during the reproduction of the dual-arm carrying motion. We observe that the robot adapts the posture of its arms to fulfill the manipulability requirements of the task. However, constraining the end-effectors to fixed positions significantly reduces the DoF redundancy that the robot can exploit to track accurately the desired manipulability. This issue may be alleviated by allowing the robot to vary the positions of its end-effector while maintaining a constant distance between them after picking up the load. 

\section{Conclusions}
\label{sec:concl}
This paper presented a detailed analysis of single and dual-arm manipulability ellipsoids for human movements during industry-like activities. Statistical analyses considering the intrinsic geometry of the manipulability ellipsoids were conducted on the kinematics data records of participants executing screwing and load carrying tasks. Our work showed that the evolution of the manipulability ellipsoid shape provides more information about human motion than the classical manipulability indices classically used in the literature. We illustrated the application of our analysis in two human-robot manipulability transfer experiments. We also showed that manipulability transfer can be utilized from a motion planning point of view, where the robot adapts its posture in anticipation to the next action before executing it.

Future work includes a data-driven automatic prioritization of the manipulability and position requirements following the approach in~\cite{Silverio19TRO}, as well as experiments on real platforms. Moreover, we will investigate how to adapt the desired manipulability to better exploit the robot capabilities. This may typically be useful to take into account the actuator constraints of the robot, as they are not considered in the human demonstrations. To do so, we may, for example, use geometry-aware Bayesian optimization~\cite{Jaquier19:GaBO}.


\bibliographystyle{IEEEtran}
\bibliography{References}

\end{document}